# Accelerating Vehicle Routing via AI-Initialized Genetic Algorithms

Ido Greenberg[1]*, Piotr Sielski[1], Hugo Linsenmaier[1], Rajesh Gandham[1],
Shie Mannor[1,2], Alex Fender[1], Gal Chechik[1], Eli Meirom[1]

**Abstract**

Vehicle Routing Problems (VRP) are an extension of the Traveling Salesperson Problem and are a fundamental NP-hard challenge in combinatorial optimization. Solving VRP in real-time at large scale has become critical in numerous applications, from growing markets like last-mile delivery to emerging use-cases like interactive logistics planning. In many applications, one has to repeatedly solve VRP instances drawn from the same distribution, yet current state-of-the-art solvers treat each instance on its own without leveraging previous examples. We introduce an optimization framework where a reinforcement learning agent is trained on prior instances and quickly generates initial solutions, which are then further optimized by a genetic algorithm. This framework, *Evolutionary Algorithm with Reinforcement Learning Initialization* (**EARLI**), consistently outperforms current state-of-the-art solvers across various time budgets. For example, EARLI handles vehicle routing with 500 locations within one second, 10x faster than current solvers for the same solution quality, enabling real-time and interactive routing at scale. EARLI can generalize to new data, as we demonstrate on real e-commerce delivery data of a previously unseen city. By combining reinforcement learning and genetic algorithms, our hybrid framework takes a step forward to closer interdisciplinary collaboration between AI and optimization communities towards real-time optimization in diverse domains.

[1]NVIDIA. [2]Technion, Israel. *e-mail: igreenberg@nvidia.com

# 1 Introduction

The Traveling Salesperson Problem [1] and its extension, the Vehicle Routing Problem (VRP) [2], are a cornerstone of combinatorial optimization with profound implications across industries like logistics [3] and urban planning [4]. In VRP, multiple vehicles with limited capacity must be assigned to delivery points while minimizing routing cost. Total U.S. freight transportation costs reached $1,391B in 2022 [5], translating every 1% improvement in routing into $10B annual saving and massively reduced carbon emissions. The theoretical and practical importance of VRP led to significant effort to develop fast and scalable algorithms for approximate solutions, such as Adaptive Large Neighborhood Search (ALNS) [6] and iterated local-search solvers [7]. In particular, Genetic Algorithms (GA) achieve many current state-of-the-art (SOTA) results [8] [9] [10]. Yet, being an NP-hard problem, VRP remains an enduring challenge for nearly 200 years [11].

VRP often presents a trade-off between computing time and solution quality. Obtaining solutions within seconds has become increasingly important for various real-world problems [3] [12] [13] [14]. These include last-mile delivery routing with on-the-fly updates, interactive “what-if” planning, and routing emergency vehicles under changing road conditions where every second may matter. In these cases, it is often preferable to "early-stop” the optimization with an approximate solution rather than optimize for a longer duration. In other applications, the time budget may vary greatly across instances, like in warehouse order picking where the optimization duration is bounded by the time until next assignment. In those cases, solvers should be capable of providing an optimized solution for any given time budget. Iterative methods like GA naturally support such early optimization stopping.

Importantly, VRP instances are often solved not in isolation but rather in a "repeated-VRP" way, where many instances from the same distribution are being optimized. For example, delivery plans within a given region may involve hundreds of related instances daily, all sharing the same roads and with customer requests from the same distribution. Popular VRP solvers like GA and ALNS do not benefit from access to data of similar instances since they solve each instance in isolation. In contrast, recent Machine Learning (ML) and Reinforcement Learning (RL) approaches leverage offline data of instances and learn to produce solutions to new instances very quickly. ML and RL approaches were applied to combinatorial optimization [15] [16] [17] [18] [19] and in particular VRP, exploiting the repeated-VRP settings [20] [21] [22] [23] [24]. Unfortunately, current ML solvers still yield lower solution quality compared to GAs, in particular in large-scale problems [25] [23]. The challenge remains how to achieve superior solutions faster by combining (a) leveraging previous instances for fast inference, and (b) enabling the user to produce solutions for any point on the trade-off curve of solution quality vs. optimization time.

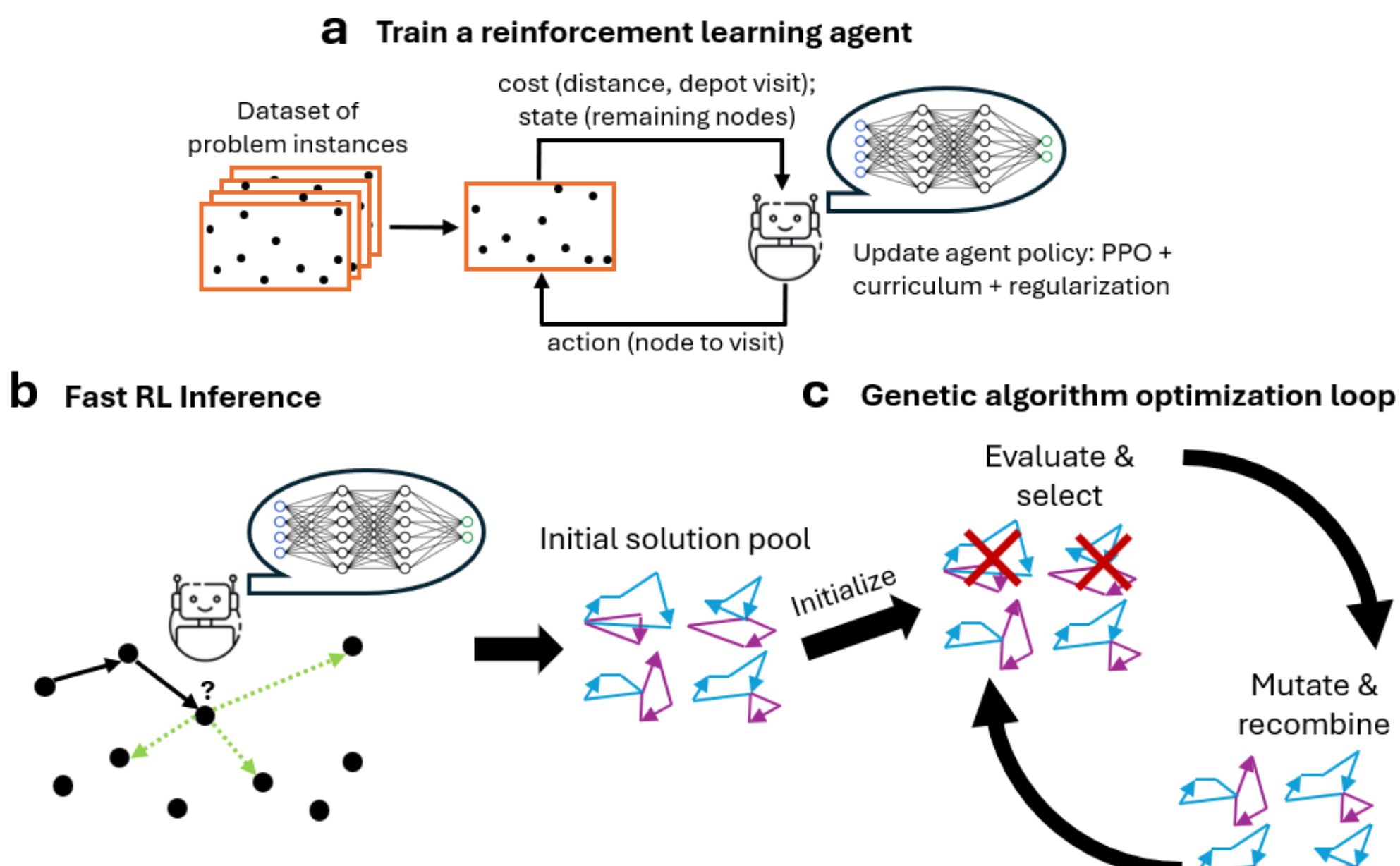

**Figure 1: EARLI: Evolutionary Algorithm with Reinforcement Learning Initialization. a,** During offline training, an RL agent interacts with a dataset of problem instances and learns to generate high-quality solutions. **b,** In production, the trained RL agent faces a new problem instance and generates $K$ solutions with quick decision making. **c,** The $K$ solutions are used as the initial population of a genetic algorithm (GA), initiating its optimization loop.

To that end, we present a framework called *Evolutionary Algorithm with Reinforcement Learning Initialization* (***EARLI***) for solving the NP-hard combinatorial optimization problem of repeated-VRP. To address the problem in real time and large scale, EARLI is designed to leverage the key strengths of its two components: the RL agent benefits from training data, and the GA provides control over the tradeoff between inference time and solution quality. The EARLI workflow is illustrated in Figure 1 and detailed in Section 4.5. After training an RL agent over a set of problem instances (Figure 1a), EARLI uses the pre-trained RL agent to quickly generate high-quality (but sub-optimal) solutions (Figure 1b); these are then used as the seed population for a GA process (Figure 1c), which improves the solutions. In this framework, RL and GA can be viewed as analogous to the two reasoning systems of the human brain highlighted by [26]. In analogy to the fast "system 1", the RL agent relies on previous experience to deliver a quick response and improves when trained with more data. In analogy to the more thorough "system 2", the GA searches for better solutions and improves when given more inference time.

While EARLI is conceptually simple, we encountered two significant challenges for generating useful initializations for a GA search. First, the number of vehicles is often the dominant objective in practical VRP, but forms a sparse, non-continuous reward signal, which is hard to optimize in RL settings. As a result, it is often ignored in RL methods which focus on distance optimization. This gap is critical in EARLI's framework: too many vehicles in the initial solutions may cause the GA to waste time or even converge to suboptimal fleet sizes. To address this challenge, we developed a capacity-aware regularization scheme that guides the RL policy toward using fewer vehicles. This outperformed a more standard approach of penalizing solutions by the number of vehicles. Second, we found that common RL approaches did not scale well on training beyond 100 customers. We scaled model fine-tuning to significantly larger problems using a curriculum learning protocol. The challenges are discussed in detail in Section 4.5.

Many methods have been proposed for initializing GAs [27] [28] [29] [30], including ML approaches [31]. EARLI's usage of offline data for acceleration differs from other initialization methods, which often aim to maximize long-term solution quality with limited effect on solver speed. For example, many ML initialization methods apply costly training at inference time without generalizing from offline instances [32] [33] [34] [35], hence do not benefit from the repeated-VRP settings. Another line of initialization methods aims to cover the solution landscape, to avoid local optima that hurt the final solution quality [28] [29]. Some learning approaches, like the Deep Ant Colony Optimization solver (DeepACO) [36], do exploit offline data in a repeated-VRP manner and generate an edge-heatmap using RL guidance. Our experiments below show that GA solvers strongly outperform this approach.

In a series of experiments, EARLI substantially improves SOTA performance on VRP with limited time-budget, and scales to hundreds of delivery points derived from both synthetic Euclidean data and real delivery data. This improvement is robust across several GA solvers, data sources, problem sizes and optimization time budgets – from sub-second up to minutes into the optimization process. In some settings, EARLI achieves in 1s the same solution quality that takes the GA over 10s to reach. Such 10x speedup can significantly enhance existing applications and even enable new use-cases of few-second optimization, for both interactive scientific research and practical applications as discussed above.

Beyond methodological advancements, this paper contributes a new benchmark for VRP, based on real-world logistics data. Conventional benchmarks often rely on synthetic data generated from unrealistic uniform distributions. Our benchmark is grounded in real e-commerce data provided by Olist [37], reflecting realistic customer locations and road-based driving durations. We further demonstrate the generalization capabilities of EARLI on real orders in a *new* city – with customers, locations and roads different from the city where the model was trained.

In summary, this work introduces a framework for accelerating routing by integrating iterative solvers with learning from experience. It provides high-quality routing solutions at speed and scale previously considered impractical. This capability can cut costs in classic industries and enable the emergence of applications like interactive logistics planning. With the release of both code and data, our framework opens new avenues for future work in both ML and optimization communities, encouraging their synergy in NP-hard optimization in general and routing problems in particular.

# 2 Results

## 2.1 EARLI accelerates time-to-solution

To evaluate EARLI in a realistic challenging scenario, we introduce a new VRP benchmark derived from e-commerce data, in addition to the standard synthetic benchmark in the literature of ML for VRP [20] [21]. The standard synthetic benchmark consists of up to 100 customers, with uniformly distributed locations and demands, and Euclidean traveling distances (as illustrated in Figure 2a). However, most real-world instances are fundamentally different: customers are often located in clusters of varying sizes and driving times vary according to the roads.

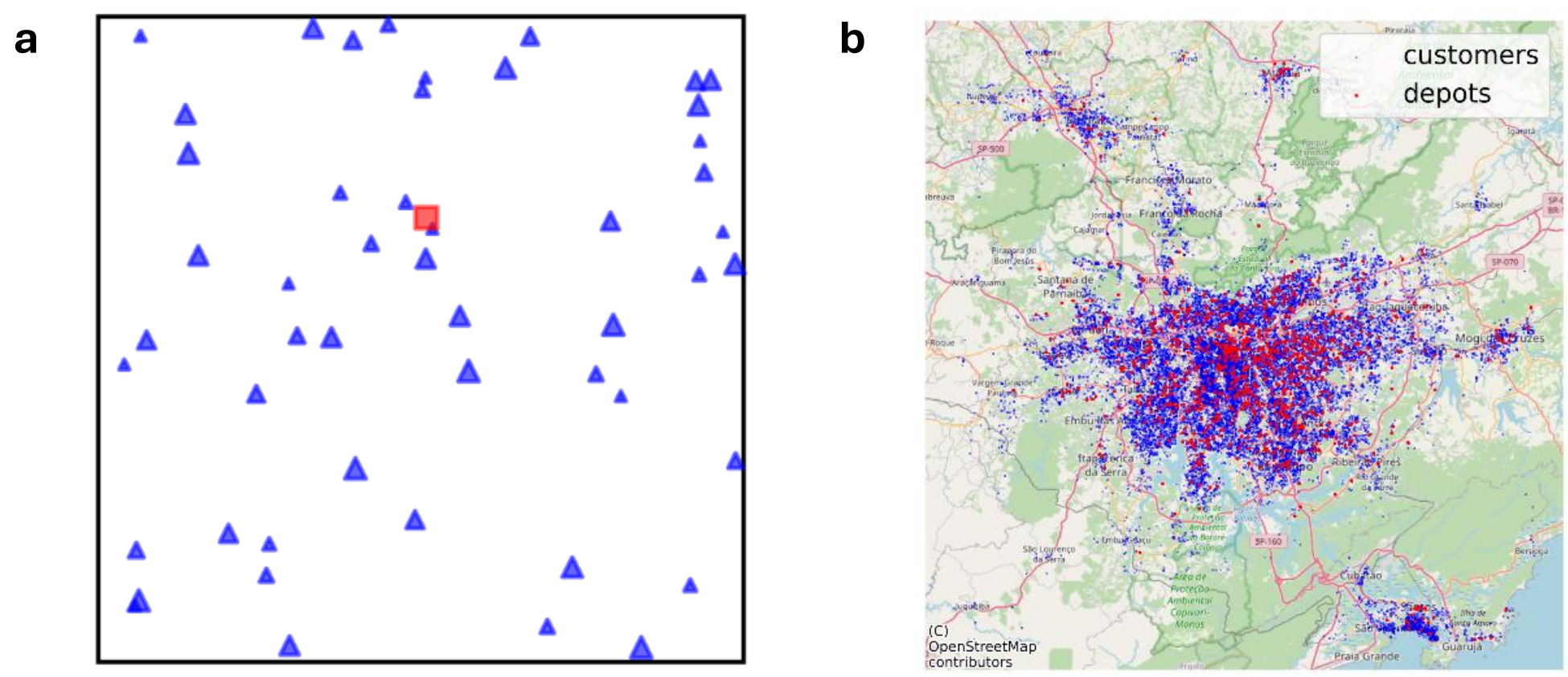

**Figure 2: Synthetic and real datasets used in this study. a, Synthetic data:** An illustration of a problem instance with 49 customers and 1 depot. Locations are uniformly distributed and travel distances are Euclidean. Random customer demands are represented by the triangle sizes. **b, Real data:** Olist orders, 100km$^2$ around Sao Paulo center, include locations of 23K customers and 1K sellers. For every problem instance, multiple customers and one depot are sampled from these locations, respectively. Travel costs correspond to driving time computed with OSRM. Demands correspond to actual product volumes.

In our real-data benchmark, the locations of customers and of the depot are sampled from a real-world dataset [37] (Figure 2b). Driving durations between locations are computed based on real roads using Project OSRM [38]. Demands are derived from real order volumes. The benchmark is described in detail in Section 4.6.

We evaluate EARLI when applied to 4 popular VRP solvers: HGS [9], cuOpt [8], PyVRP [39] and LKH3 [7]. The first three are based on GAs, and LKH3 relies on an iterative local-search operator. HGS and cuOpt report SOTA results on various routing problems [9] [8]. By default, all 4 solvers initialize their population with random solutions. We test each solver with its own random initialization, with our proposed RL initialization, and with various alternative initialization methods specified below. For each of the 4 solvers, we compare initialization schemes over 256 test instances across a range of time budgets. For every time budget, we show the gap between the obtained solution cost and the best-known solution cost, defined as the lowest cost amongst all solvers, initialization schemes and time budgets.

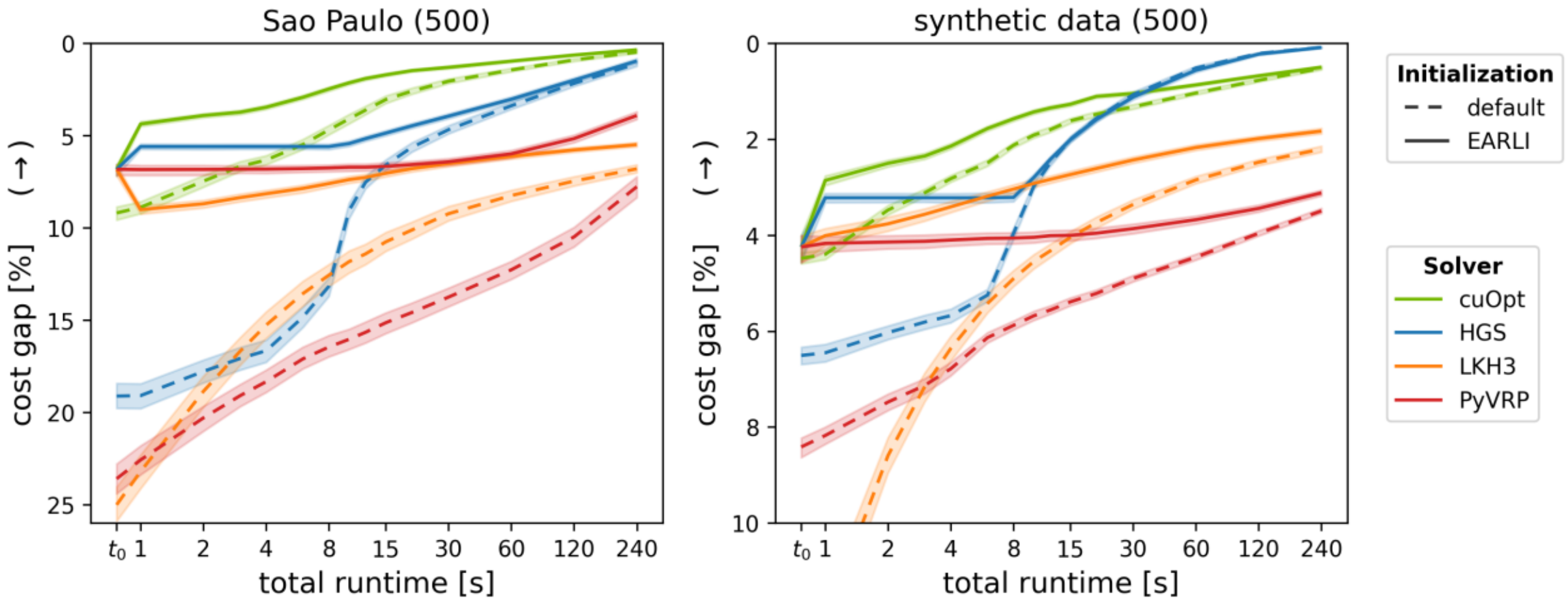

**Figure 3: EARLI improves solution quality given a fixed time-budget**. EARLI (solid lines) improves the mean cost across a variety of optimization times from seconds up to minutes. Cost gaps are averaged over 256 test instances of 500 customers. Shading corresponds to 95% confidence intervals. $t_0$ corresponds to the runtime of the RL initialization on its own, before applying the GA.

Overall, as presented in Figure 3, **EARLI obtains state-of-the-art solution costs across a wide range of time budgets in both real data and synthetic data benchmarks, improving all 4 solvers up to minutes into the optimization process.**

EARLI combines an RL agent with a GA solver and strongly outperforms each of the two on their own. For example, compared to the RL agent alone (time $t_0$ in Figure 3), EARLI with HGS improves solution quality almost immediately. **Compared to HGS alone with time budget of 6s, EARLI improves solution quality by 7.7% on real data**. **In fact, it achieves within 1s the same average solution quality that takes HGS over 10s to reach with its default initialization, obtaining 10x optimization speedup in this scenario.**

Our experiments focus on problem sizes of 100-500 customers, a regime that poses a significant challenge for existing solvers given time budget limited to seconds. Figure 4 demonstrates that the advantage of EARLI increases with the problem size, as larger instances pose a harder challenge for the GA solver.

We further compared EARLI to several alternative initialization methods: (a) the **default** random initialization; (b) a greedy **nearest-neighbor** initialization procedure that was previously shown effective for routing problems [40]; (c) **TNT** and (d) **QBL**, two metaheuristics that encourage uniform coverage of the solution space, and performed best among the candidates evaluated in [28]. We further consider (e) **DeepACO** [36], which also uses RL to learn from data and guide an Ant Colony Optimization solver. In contrast to the other methods, DeepACO is a complete pipeline that includes its own ACO solver rather than a GA initialization scheme. We test DeepACO following the settings of [36], namely, synthetic problem instances with 100 or 500 customers. In the supplemental, we also test the approach of (f) test-time AI-based optimization [32], demonstrating that learning each problem by itself provides significantly slower optimization. As shown in Figure 4, EARLI provides significant value over all compared initialization methods. Detailed cost figures for 100, 200 and 500 customers are available in the Extended Data.

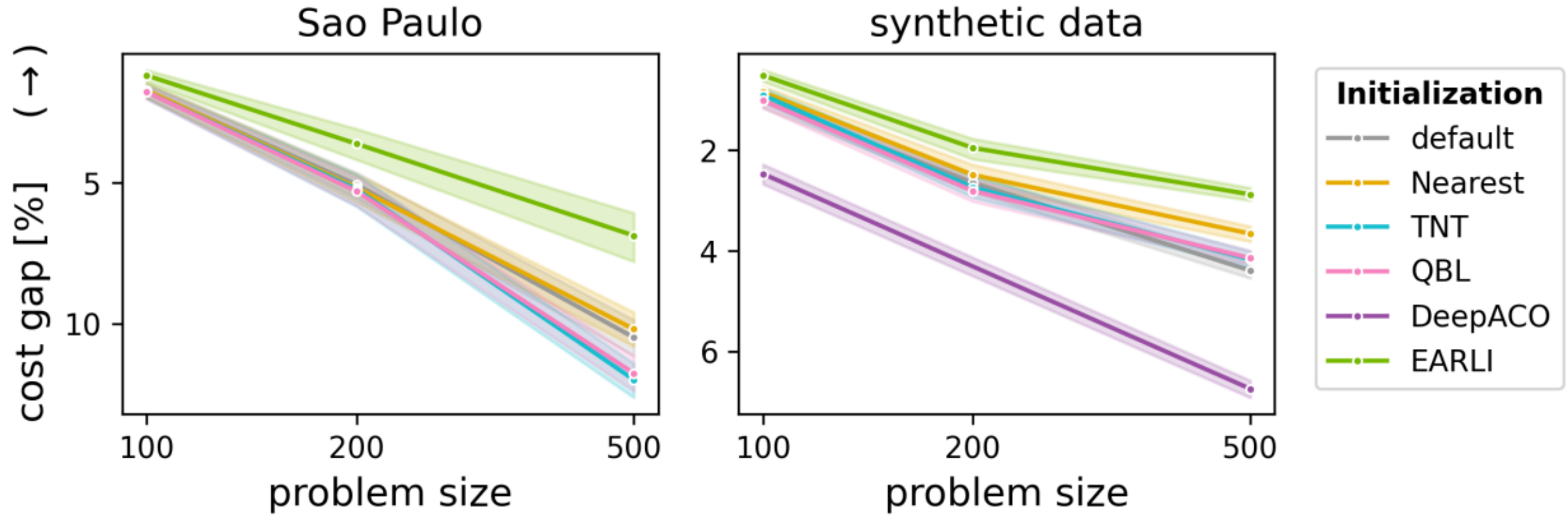

**Figure 4: EARLI outperforms alternative initializations, and its benefit grows with problem size.** The solver (cuOpt or DeepACO) is given a time budget of 1s per problem. Shading corresponds to 95% confidence intervals over 256 test instances.

Figure 3 and Figure 4 reveal how EARLI improves solution costs in instances where feasible solutions are found, but for some instances a feasible solution may not be found within the allotted time. Note that vehicle minimization coincides with feasibility in our settings, as the vehicle budget is tight (see Section 4.6). To evaluate *both* the feasibility and the cost of solutions together, we measure the improvement of EARLI on each problem instance via a hierarchical objective: first prioritizing feasibility, then the cost. That is, EARLI wins on a problem instance if it finds a feasible solution while the GA alone fails; or if both find a feasible solution, and the cost of EARLI is lower; and vice versa. Perfect ties (3% of the instances) are excluded. Figure 5a presents the percentile of EARLI wins. EARLI improves the result in over 85% of the instances in the few-second regime. This is also evident in Figure 5b, presenting EARLI's cost advantage over the vanilla GA, in each one of the 256 test instances.

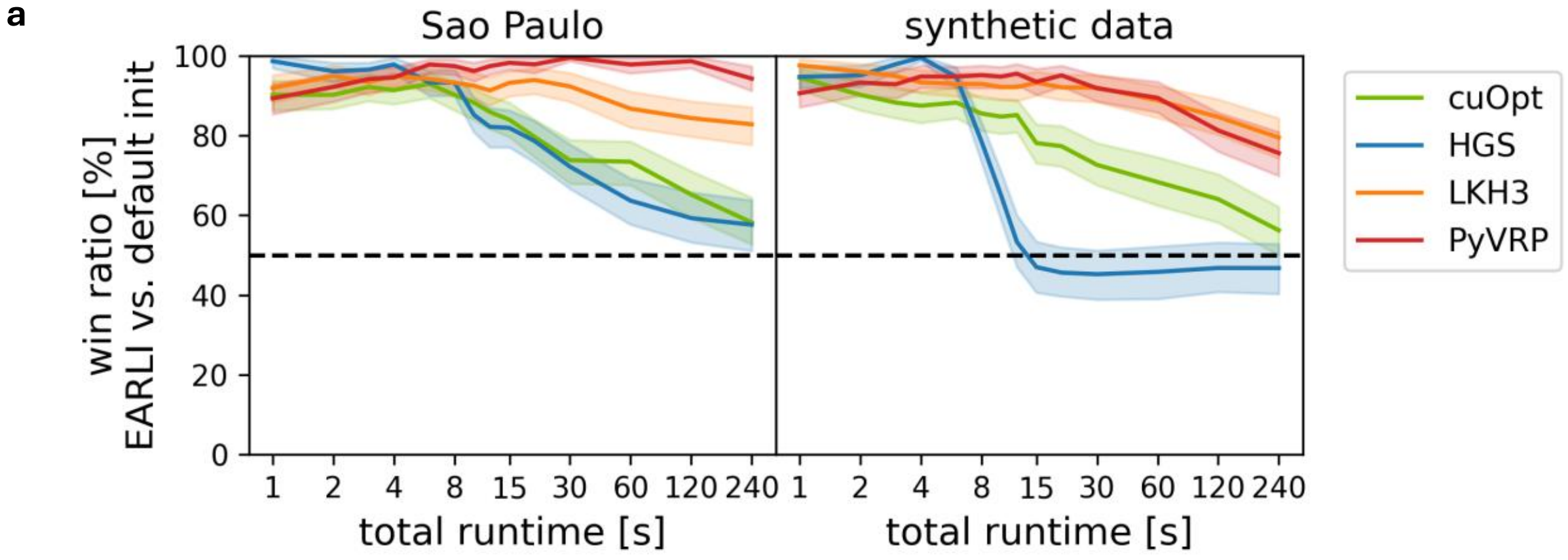

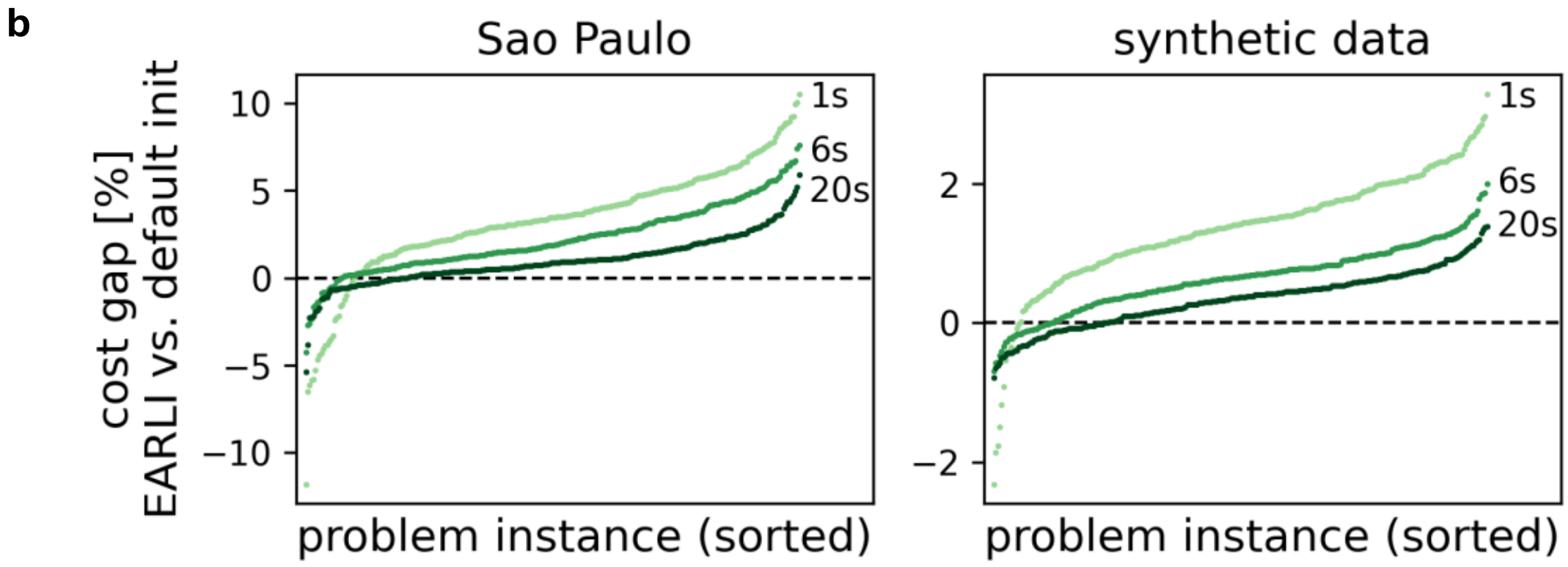

**Figure 5: a, Win ratio:** the percentage of problem instances where EARLI is superior to the default initialization. A solver's win is defined as finding a feasible solution where its competitor fails; or obtaining a better cost, if both found a feasible solution. Percent is calculated out of decisive problem instances (ties are excluded). Shading corresponds to 95% confidence intervals. **b, Cost reduction of EARLI** for cuOpt solver, for each of the 256 test instances with 500 customers, for time-budgets 1s, 6s, 20s.

## 2.2 Robustness to domain shift

A key requirement in real world applications of VRP is that trained models would generalize outside their training distribution. For example, once a model is trained to optimize VRP instances in one set of cities, one may want to apply the same model to new cities without conducting expensive data collection and training.

In this section, we test the robustness of EARLI to a shift in the distribution of problem instances. We begin by transforming synthetic test instances. We avoid linear transformations like rotation and translation, which are common in routing but would preserve the uniform customer distribution. Instead, we apply Barrel and Pincushion transformations (Figure 6), commonly used in computer vision [41] [42], to create significant distortions from the uniform customer distribution on which EARLI was trained. In terms of city topologies, Barrel resembles ring-like structures like outer boroughs around a city core, and Pincushion resembles a dense center with radial extensions. EARLI never observed a warped problem instance throughout training, yet its advantage over alternative initialization methods is preserved in these deformed test instances. This demonstrates EARLI's robustness and potential to leverage a single train dataset for different applicational domains.

To further test the robustness of EARLI to real-world distribution shifts, we train the RL agent on real data from Sao Paulo and test it on routing instances in Rio de Janeiro. The test dataset is sampled from a different distribution of customer locations, and an entirely different road layout, with the huge impassable Guanabara Bay near its center. Figure 7 shows that EARLI strongly reduces solution costs on test data despite the distribution shift. Figure 8 displays a sample solution of cuOpt, with and without EARLI.

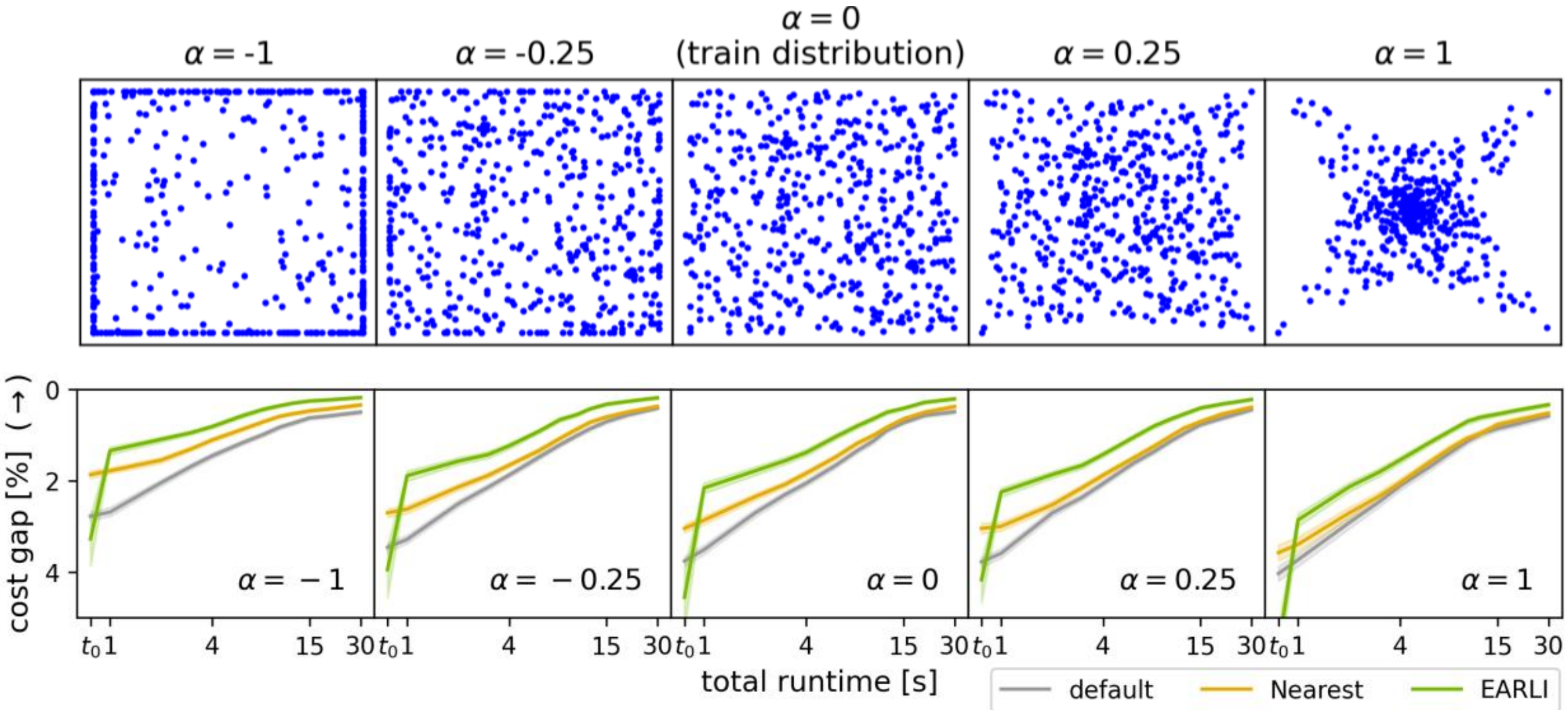

**Figure 6: Distribution shifts between train and test instances.** Top: a sample test-problem with varying distortion level. $\alpha = 0$ corresponds to no distortion. Bottom: cost gaps of the cuOpt solver with different initialization methods, with 95% confidence intervals over 256 test instances of 500 customers.

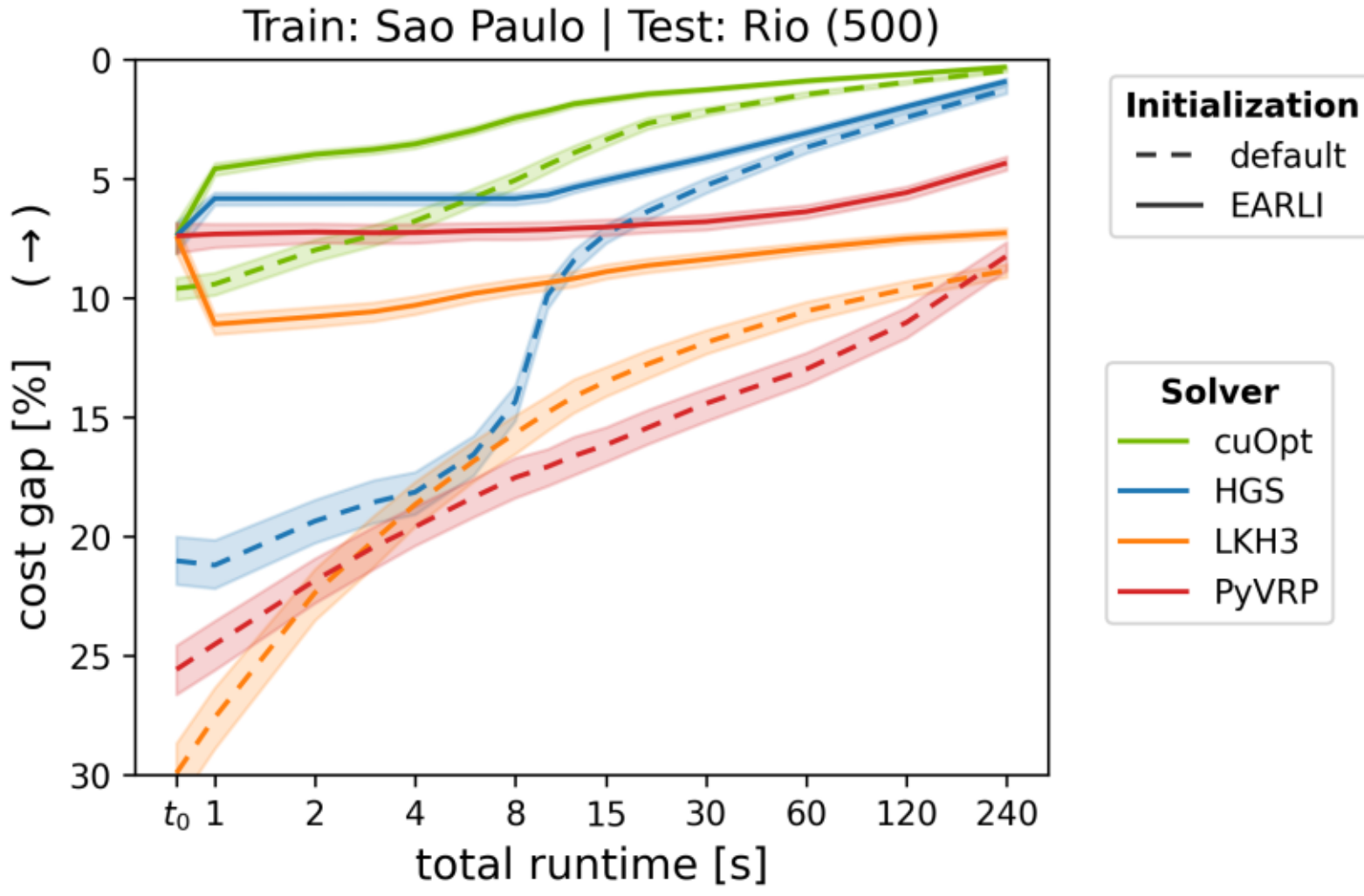

**Figure 7: Cost gaps under domain shift to a new city**. EARLI's RL agent is trained on Sao Paulo data but is tested in Rio de Janeiro. Mean gaps and 95% confidence intervals are shown over 256 test instances with 500 customers.

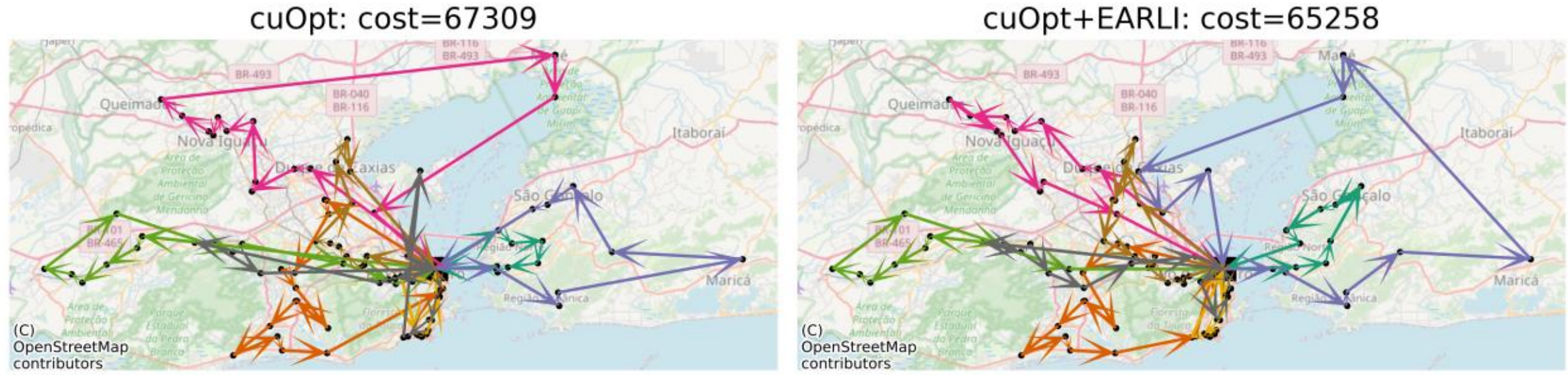

**Figure 8:** Solution obtained with cuOpt with and without EARLI in a sample problem of 100 customers in Rio de Janeiro, given a time-budget of 1s. The RL agent was trained on Sao Paulo instances and generalized to Rio. Arrows are straight for visualization only: the actual traveling costs correspond to road-based driving time. More examples are provided in the Extended Data.

Finally, we evaluate how the size of the training dataset impacts the solver. This is important, for instance, if only a handful of training examples are available, or if the underlying distribution is hard to sample from. To answer this, we consider train datasets with varying numbers of instances and train an RL agent for each dataset size. Every agent is trained for the same number of iterations, and a smaller dataset means that each problem instance is observed more frequently. To enable multiple full training runs in this experiment, we used smaller instances with 100 customers. As presented in Figure 9, EARLI provides significant acceleration even when trained on as few as 40 instances and seems to exhaust most of the value of the data at several hundred instances.

In summary, we find that EARLI demonstrates low sensitivity to data quantity (number of problem instances) and to distribution shifts that we tested.

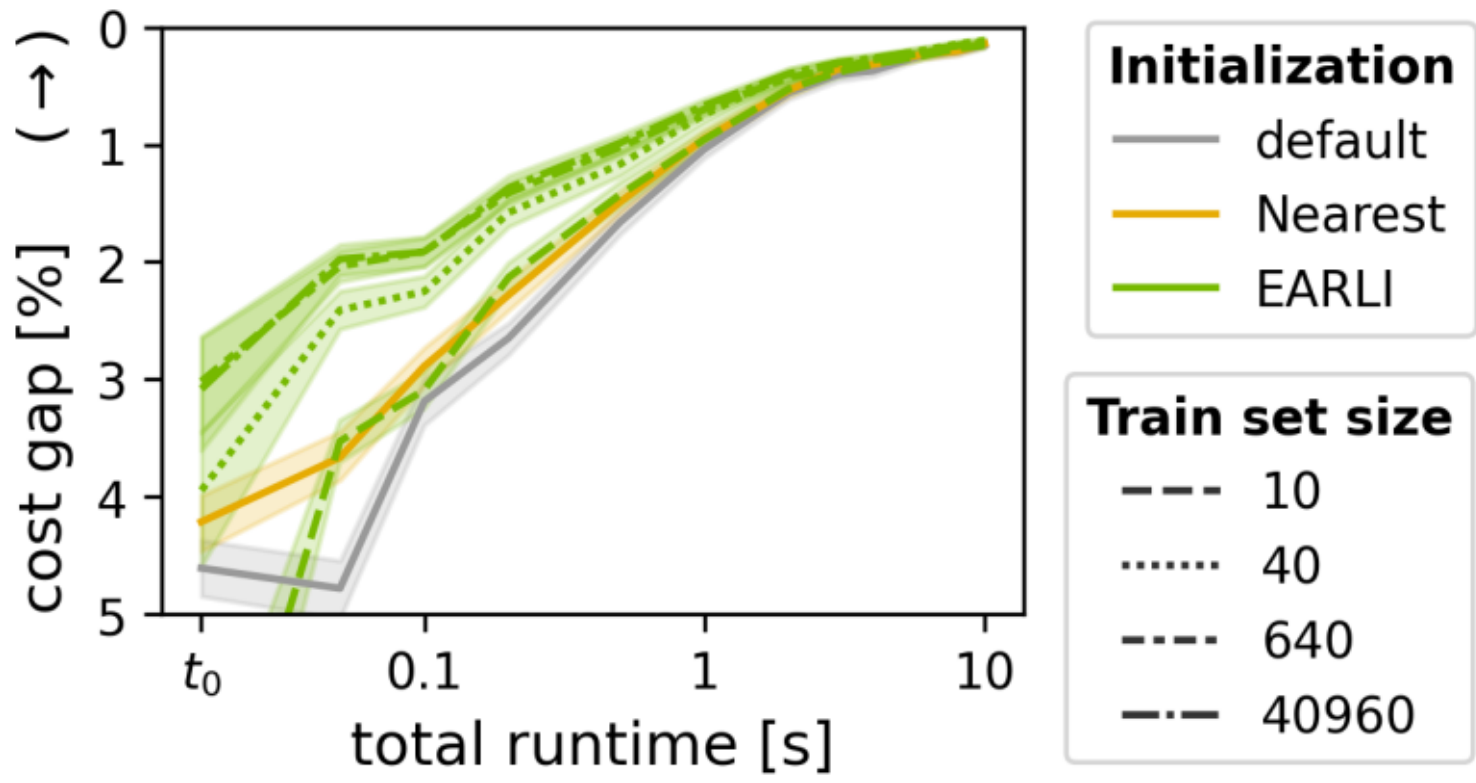

**Figure 9: Sensitivity to the size of training data set.** Cost gaps are shown for cuOpt when initialized by different RL agents trained on different number of train instances. Showing mean gaps with 95% confidence intervals over 256 test instances of 100 customers.

# 3 Discussion

We put forward an optimization setup of “repeated VRP”, where an optimizer has to solve many instances drawn from the same distribution. We show that this real-life setup can greatly benefit by combining learning-based methods, and specifically RL, with state-of-the-art optimization techniques based on GAs. In this section we discuss some implications of the experiments presented in Section 2.

**The importance of synergy between learning and optimization**

The results of our study highlight the potential in the synergy between machine learning and traditional optimization approaches. By combining the strengths of RL's rapid decision making with GA's thorough search capabilities, EARLI achieves performance that surpasses what either method can accomplish alone. While these fields have often developed in parallel, in separate communities, our results demonstrate the value of inter-disciplinary research that carefully acknowledges the benefits of each approach and integrates them accordingly. As both fields continue to advance, we anticipate that cooperative efforts will become increasingly crucial in tackling real-world challenges that require both data-driven insights and sophisticated optimization techniques.

**The importance of initialization**

The results indicate that typical GAs spend a long time on finding a "reasonable" solution from which the optimization goes on. This long warm-up time is what allows smart initialization to significantly accelerate the optimization process. Our novel RL initialization approach provides the greatest value in our experiments. Surprisingly, even a simple initialization with nearest neighbor solutions sometimes provides moderate acceleration over SOTA solvers (as shown in Figure 4 and more extensively in the Extended Data). Yet, and despite the extensive literature about GA initialization (discussed in Section 4.4), SOTA VRP solvers use random initialization by default, and often do not permit an initialization interface at all. For HGS and PyVRP, for example, we developed and open-sourced a dedicated initialization interface.

The preference of a pure random initialization may be motivated by bias prevention: there may be a concern of greedy solutions biasing the solver towards local optima. Indeed, as discussed in Section 4.4, much of the GA initialization literature focuses not on high-quality initial solutions, but rather on covering the search space. Still, no bias was observed in our experiments, and an initialization interface may empower the user to initialize the solver according to the problem and to their needs. The initialization may express experience from data (as in this work), a systematic cover of the search space, or any other domain knowledge of the user.

**The importance of robustness and generalization**

Our experiments demonstrate the **robustness** of the proposed method across different problem sizes (100-500 customers); distributions (synthetic and real-world data); time scales (from sub-second to a few minutes); train dataset sizes (from tens to thousands of instances); and solvers (cuOpt, HGS, LKH3 and PyVRP). Particularly noteworthy is the **generalization** beyond the settings of the train data, as the RL agent trained on Sao Paulo data was still able to accelerate the optimization of routes in Rio de Janeiro. Both robustness and generalization capabilities are crucial for real-world applications, where the ability to adapt to new cities or changing distribution patterns is often necessary.

In conclusion, our work demonstrates the significant benefit of combining machine learning techniques with traditional optimization methods for solving complex combinatorial problems. By leveraging the strengths of both approaches, we have developed a method that not only improves solution quality in short time budgets but also shows promise for generalization and scalability. While this work focuses on the VRP, the approach of EARLI is applicable to other combinatorial optimization problems as well and takes a step forward in the research of NP-hard optimization.

As the fields of artificial intelligence and operations research continue to grow, we anticipate that hybrid approaches like EARLI will play an increasingly significant role in solving real-world optimization challenges.

# 4 Methods and Literature

## 4.1 The Capacitated Vehicle Routing Problem (VRP)

The Capacitated Vehicle Routing Problem (VRP or CVRP) is a fundamental challenge in discrete optimization. This NP-hard problem generalizes the Traveling Salesperson Problem (TSP) and forms the basis for many real-world applications in logistics and transportation.

A VRP instance is formally defined over a graph $G = (V, E)$, where the vertices $V$ represent a depot and delivery points, and the edges $E$ represent travel costs (e.g., distances). Each delivery $i$ is associated with a known demand $d_i$. A fleet of $K$ vehicles with capacity $Q$ is stationed at the depot (vertex $i = 0$). The problem instance is fully defined by $I = (G, \{d_i\}, K, Q)$. A solution is defined by a partition that assigns each vertex to a vehicle route, and by the order of deliveries within each route. In a feasible solution, all routes start and end at the depot, each customer is visited exactly once, and the total demand in each route does not exceed $Q$.

The goal is to find a feasible solution that minimizes the objective, which has two different versions in the literature: (1) minimize the total traveling cost over all the routes [21] [7]; or (2) a hierarchical objective, first minimizing the number of used vehicles (number of routes), then minimizing the total cost over them [8] [43]. In this work, different solvers may optimize different objectives. To guarantee that the two objectives coincide and that all solvers are compared fairly, we fixed the vehicle constraint $K$ to the minimal known number of vehicles required to solve the problem. As a result, all optimizers had the same number of vehicles and feasible solutions only differ by the traveling cost.

We define the *repeated-VRP* problem, where instances are drawn from a fixed distribution $I \sim D$. The distribution $D$ may express recurring structure, such as distances derived from the same road network, or consistent demand scales and vehicle capacities. $D$ is unknown but can be sampled offline for training. During inference on new instances, the solver may leverage pre-learned knowledge without reducing its time budget.

## 4.2 VRP Solvers

Genetic Algorithms (GAs) have been widely applied to solve Vehicle Routing Problems due to their ability to effectively explore large solution spaces, often achieving SOTA results [9] [8] [10]. In the context of VRP, a GA typically represents solutions as chromosomes encoding parts of vehicle routes (edges in the problem's graph representation). The algorithm evolves a population of solutions through selection, crossover, and mutation operators specifically designed for routing problems. Common crossover methods for VRP include order crossover and edge recombination crossover, while mutation operators often involve local search moves like 2-opt or node insertion. GAs for VRP often find high-quality solutions, especially when hybridized with local search techniques.

In this work, we evaluated EARLI with several top-tier VRP solvers:

**CuOpt** [8]: cuOpt is a general-purpose vehicle routing solver covering numerous variants of the problem including time windows, waiting times, precedence constraints, pickups and deliveries, prize collection, heterogeneous fleet, and multiple depots, among others. The core of the framework consists of evolutionary algorithms, advanced diversity management, fast local search, approximate search, infeasibility exploration and large neighborhood search.

**HGS** (Hybrid Genetic Search [9] [44]): HGS is a powerful metaheuristic that combines the global exploration capabilities of genetic algorithms with the intensification strength of local search. It uses efficient solution representation, advanced crossover operators, and a diversity management scheme to maintain a balance between solution quality and population diversity.

**PyVRP** [39]: This is an open-source implementation of hybrid genetic search for VRP. PyVRP is designed to be easily extensible and customizable, allowing researchers to build upon a SOTA solver. It combines the flexibility of Python with the performance of C++ by implementing critical parts of the algorithm in C++.

**LKH3** (Lin-Kernighan-Helsgaun [7]): LKH3 solver is based on an iterative local-search operator that reaches exceptional performance on VRP instances. It is an extension of LKH [45] for the Traveling Salesperson Problem (TSP), adapted to handle the various constraints in VRP. LKH3 uses sophisticated local search techniques and has shown remarkable results on many VRP benchmark instances.

Various alternative methods have been proposed beyond GAs to solve VRP. OR-Tools [46] combines classical heuristics with constraint programming. Adaptive Large Neighborhood Search (ALNS) [6] dynamically selects among destroy-and-repair operators; and in particular, Slack Induction by String Removals (SISR) [47] applies a large-neighborhood search based on removing adjacent customer strings and reinserting them.

## 4.3 Reinforcement Learning for VRP

Machine learning (ML) approaches, and reinforcement learning (RL) in particular, have emerged as promising methods for solving the Vehicle Routing Problem (VRP) in recent years [20] [21] [22]. These techniques offer a fundamentally different approach compared to classical optimization methods. Rather than solving each instance from scratch, ML approaches aim to learn policies or heuristics that can generalize across different problem instances. This is typically done by training models on numerous VRP instances drawn from the same distribution.

ML approaches do not rely on problem-specific knowledge or hand-crafted rules and can potentially discover novel strategies and adapt to different problem variants without significant human intervention. They also can leverage patterns and structures in the data that may not be apparent to human designers.

In supervised ML approaches, the solver learns to mimic existing solutions (e.g., provided by classic solvers). By contrast, RL aims to find a solver policy that minimizes solution costs, hence does not depend on external solvers and is not limited by their solution quality. In an RL formulation, VRP is framed as a sequential decision-making problem: the agent learns to construct solutions by making a series of decisions (e.g., which customer to visit next) based on the current state of the problem (e.g., current vehicle location and capacity, remaining unserved customers, and their demands). Recent works have explored various types of supervision for VRP, as well as different modeling architectures, including graph neural networks and attention mechanisms [20] [21].

**Learning-from and scaling-with data**

A key advantage of ML is its ability to learn from data, instead of requiring hand-crafted heuristics. This allows ML to scale its quality with the amount of data, which is particularly beneficial as data availability becomes higher, or whenever problem instances can be easily simulated. Once learning is complete, the learned experience allows the ML solver to present a significantly faster inference time.

**Limited improvement with running time**

Unfortunately, ML solvers have struggled to compete with GA solvers so far, and provided sub-optimal solutions [21] [22]. A common challenge in ML approaches is to effectively exploit inference time: once presenting a solution, many ML methods are incapable of improving it directly; even if more running time is allocated, the ML solver will often generate additional solutions instead of improving the existing ones, limiting the cost improvement [21]. We tackle this limitation by using the ML solutions as a warm-start for the GA solver, which lets the solutions improve gradually as more inference runtime is permitted.

**Generalizing to realistic instances**

Another limitation of ML is its sensitivity to the data. In the literature of ML for VRP, solvers are typically trained on a quite specific distribution of instances, often with up to 100 customers, uniformly-random locations and Euclidean distances [20] [21]. While the literature presents impressive results for this setting, the learned solvers are usually not tested against more realistic instances. Here, we tackle this limitation by presenting a novel experimental benchmark derived from real delivery data, as presented in detail in Section 4.6.

## 4.4 Initialization of Genetic Algorithms

Traditionally, random initialization has been the most common method for populating the initial generation in GAs; yet, while simple and unbiased, this approach often leads to slower convergence and suboptimal solutions [27] [28]. Recognizing this limitation, various initialization alternatives were explored. Some studies focus on covering the solution space, e.g., using Low Discrepancy methods to minimize non-uniformity [28], or Opposition-Based initialization to include each initial solution along with its “opposite” [29]. Quasi-opposition Based Learning (QBL) [48] and Tent mapping (TNT) [49] were found to provide the best improvement for function optimization problems, among the initialization methods evaluated in [28]. However, due to the curse of dimensionality, uniformly covering the solution space is often impractical in large-scale problem instances, and high-quality initial solutions become the key to acceleration of the GA [27] [28].

To improve initial solution quality, various studies propose greedy or manually crafted initialization methods, such as gene-banks [50] and KNN subgraphs [30]. In the greedy nearest neighbor approach, the next customer to visit is always the closest one to the current customer, among remaining feasible customers. Such approaches are sometimes combined along with random solutions, referred to as hybrid initialization [51]. Many initialization methods are inevitably tailored for the specific problem at hand, e.g., TSP [30], feature selection [52] or the P-median problem [53]. Still, so far, the greedy nearest neighbor solver remains a competitive candidate for quality-based GA initialization [40]. EARLI provides a substantial improvement beyond the nearest neighbor method, as presented in Section 2 and in the Extended Data.

Recent works leverage various ML tools to enhance traditional solvers like GAs. RL was proposed to learn better GA operators [54] [55]. Test-time ML training was applied to assist iterative solvers in the beginning [32] [33] [35] or throughout the search process [34]. Such test-time training leverages ML tools to optimize a given problem instance, but does not learn from previous instances. In the supplemental, we demonstrate that this approach requires significantly longer running time on inference compared to learning from offline training.

## 4.5 EARLI: Challenges and implementation

EARLI leverages offline data of instances to learn how to generate a high-quality initial population of solutions for a GA solver. The RL agent generates a solution step by step: at each step, it observes the problem instance, the set of already-visited vertices, and the remaining vehicle capacity, and decides which vertex to visit next. The reward for training is the travel cost to the next vertex, with additional regularization as discussed below. We train a policy based on an Attention Model similar to [56]. The policy outputs a distribution over the feasible actions, i.e., the remaining nodes to be visited. A stochastic solution is generated by iteratively sampling this distribution, and a deterministic solution by choosing the action with highest probability.

We optimize the policy using the popular PPO algorithm [57]: at each train epoch, the current stochastic agent runs on a batch of instances; measures the success of each action according to the following trajectory cost, compared to the expected cost from the same position; and updates the model accordingly, before running the next batch. The expected cost is estimated by a learned value function, reduces variance and accelerates training.

At inference time, the trained RL agent generates a batch of solution candidates, which are refined via local search and used to initialize the iterative solver. This typically provides an improved starting point for the solver with little overhead. Full details are provided in the supplemental.

Below we discuss key challenges of EARLI and the strategies we used to overcome them.

### Scalable RL training via curriculum learning

Scaling RL to large VRPs is often unstable and prohibitively slow in practice: direct training on 500-customer instances barely improved solution costs after days of training, with remaining cost gaps of hundreds of percent. Standard RL techniques for variance reduction in long horizons, like Generalized Advantage Estimation (GAE), did not stabilize training. We therefore adopt a variant of curriculum learning [58]: train on 50-node instances and gradually increase size, fine-tuning at each stage on new larger instances. This enables stable learning at tractable training times. In addition, unlike alternative curriculums for VRP [59], our method naturally produces a family of size-specific models that can be used for different problems, as demonstrated in our experiments.

### Vehicle minimization via capacity regularization

Most RL works for VRP focus on minimizing cost or distance, ignoring the number of vehicles [20] [21]. By contrast, many practical problems prioritize vehicle minimization, as supported by common solvers [8]. We found empirically that initial solutions with sub-optimal vehicles often slow down or even prevent the solver from optimizing the vehicles. We address this by (a) encourage the RL agent to minimize the number of vehicles; (b) only feed the solver with initial solutions whose number of vehicles is *guaranteed* to be optimal, as detailed below.

(a) To encourage the RL agent to minimize vehicles, we first attempted straight-forward vehicle penalty on training: every time the agent adds a new vehicle, a fixed cost is added. We found this approach limited, especially on large problem instances. We hypothesize the feedback is too delayed to be learned. For example, consider the last order of the first vehicle in a solution: if its demand does not utilize the remaining vehicle capacity, the immediate vehicle penalty remains the same, yet the capacity waste may crucially lead to an additional vehicle in the end. This extra penalty is only observed hundreds of steps later.

Instead, we propose a novel capacity regularization penalty. At the end of each vehicle route, we add a penalty proportional to the remaining capacity in the vehicle:

$$penalty(vehicle\ route) = \lambda\left(Q - \sum_{i \in route} d_i\right),$$

where $Q$ is the vehicle capacity, $d_i$ are the demands served in the current route, and $\lambda$ is the scale coefficient of the regularization. This provides immediate feedback on capacity waste, allowing the RL agent to strategize vehicle minimization without propagating penalties throughout hundreds of steps.

To test our capacity regularization method, we take an RL agent pre-trained for 200 customers in Sao Paulo, and fine-tune for 500 customers. For each regularization method, fine-tuning has a tradeoff between vehicle minimization and net cost minimization, depending on the scale coefficient $\lambda$. As shown in Figure 10, capacity regularization produces a superior trade-off curve compared to direct vehicle regularization. In addition, vehicle regularization provides only limited control over the trade-off, as increasing $\lambda$ does not necessarily improve vehicle minimization.

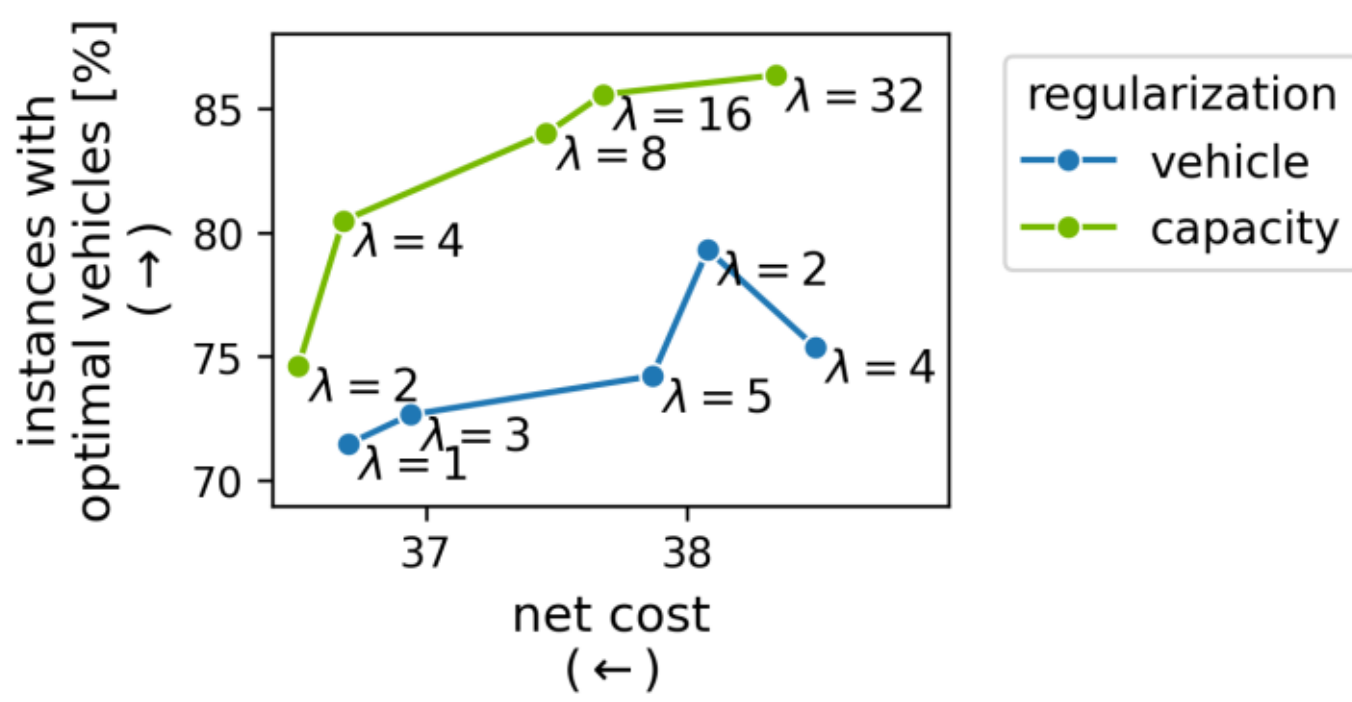

**Figure 10: Capacity regularization improves the tradeoff between vehicle minimization and net cost minimization.** Every point represents fine-tuning of the RL agent with a different regularization scale $\lambda$, tested over 256 instances with 500 customers.

(b) To verify that the solver is fed only solutions with an optimal number of vehicles, we recall that the minimal number of vehicles for a VRP instance is lower bounded by $\left\lceil \frac{total\ demand}{capacity} \right\rceil$. In the experiments, our RL solutions have met this lower bound in 87-96% of the problem instances (depending on the benchmark, as displayed in the Extended Data). Hence, in all these instances, at least one RL solution was *guaranteed* to obtain the optimal number of vehicles. We fed EARLI's initial solutions to the GA solver *only* in these cases. In the remaining instances, we used instead the greedy nearest-neighbor solution for initialization; and if it did not meet the lower bound either, we simply executed the solver without initialization, with the remaining time-budget.

**Solution diversity**

To enforce the RL agent to generate diverse solutions, we attempted penalizing choice of the same edges in different solutions. As the agent generates multiple solutions in parallel, we reduced the selection probability of edges that were already selected by other solutions. However, this resulted in quality degradation with only moderate increase in diversity and did not improve the final solution quality. Thus, we turned to a simpler approach relying on the learned RL policy distribution to sample a diverse set of solutions, as detailed in the supplemental. We leave further research on solution diversity for future work.

## 4.6 Experimental details

As mentioned above, in our experiments we (a) use the standard synthetic benchmark for 100 customers; (b) extend it to larger problem sizes; and (c) introduce a novel real-data benchmark.

**Synthetic data**

For synthetic instances of 100 customers (or more precisely, 1 depot and 99 customers), we use the same problem configuration as in [21]: i.i.d uniformly distributed locations in a square; Euclidean distances; uniformly distributed demands in $\{1,2,\dots,9\}$; and vehicle capacity of 50. We extend this setting for 200 and 500 customers, where we set the vehicle capacity to vary randomly per problem instance, uniformly in $\{40,41,\dots,80\}$. Notice that a varying capacity is mathematically equivalent to a varying scale of demands, as motivated by different types of deliveries. The average capacity (60 for 200-500 customers vs. 50 for 100 customers) respects the convention in the literature, where capacity increases with problem size. For the extended synthetic benchmarks of 200-500 customers, we train an RL agent as discussed above. For the original benchmark of 100 customers, we use the trained RL agent published by [21].

**Real data**

The real data used in our study is derived from the "Brazilian E-Commerce Public Dataset by Olist" [37], which encompasses 100,000 orders placed in Brazil between 2016 and 2018. This dataset originates from the Olist Store, an online platform that connects buyers and sellers (similar to services such as eBay and AliExpress). We focus on two subsets of data, each within a 100km$^2$ area, centered around Rio de Janeiro (8,758 orders) and Sao Paulo (23,197 orders). For each of the two, we separate the dataset into 3 ranges of dates, intended for training, validation, and test problem instances.

To maintain customer privacy, each order's location is only specified by a zip code. A separate data table specifies a list of coordinates per zip code, from which we randomly draw a specific location to associate with each order. 2% of the resulting samples corresponded to duplicated locations, which were removed from the data.

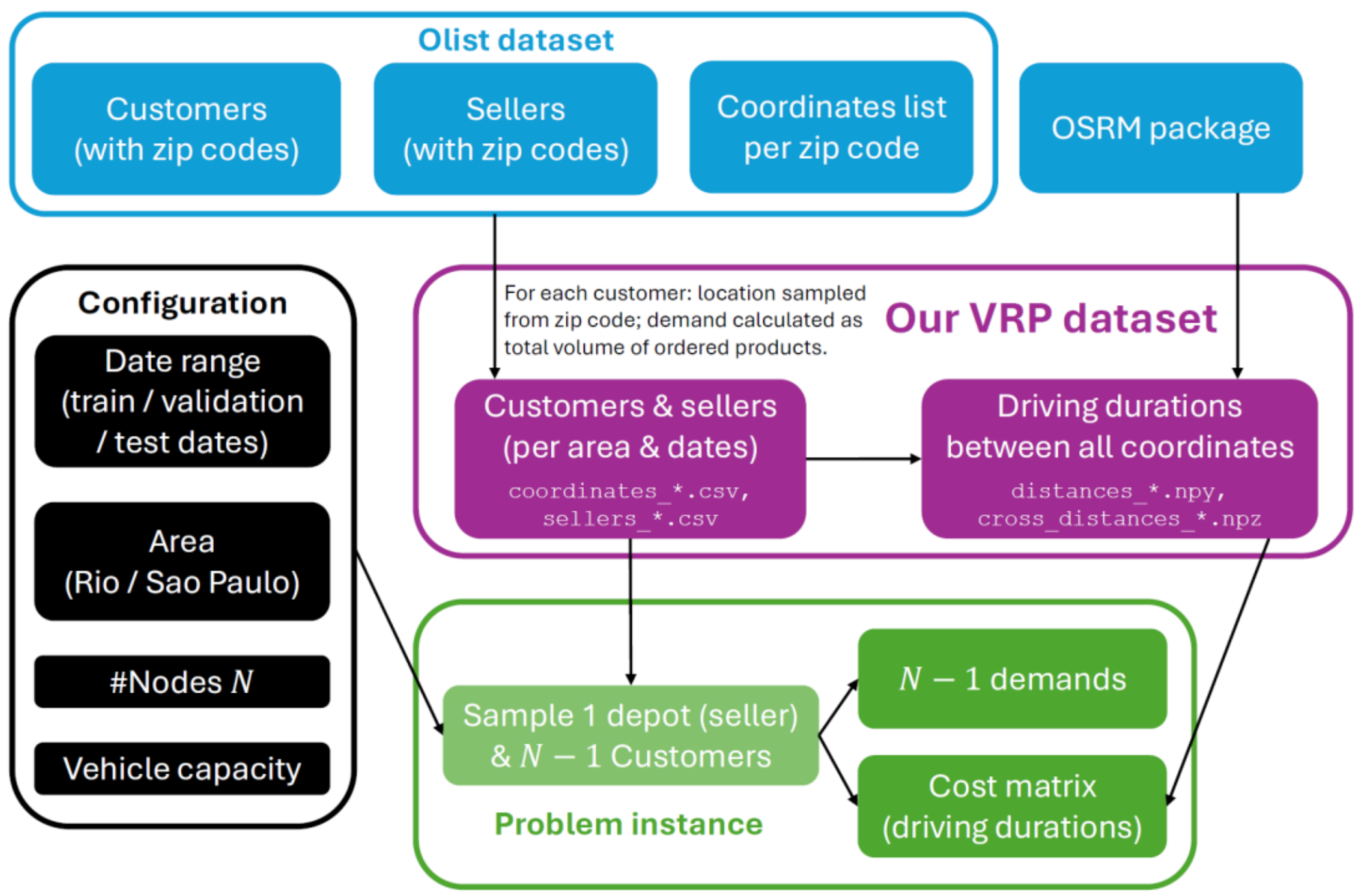

**Figure 11: Real-data benchmark for VRP.** The benchmark relies on the orders data by Olist and driving times calculated by OSRM. Problem instances are sampled according to the desired region and problem size.

To generate a new problem instance with $N-1$ customers, we simply draw $N-1$ random coordinates from the list of orders in the selected area and date-range. The location of the $N^{\text{th}}$ node – the depot – is sampled similarly from the list of sellers (out of 136 sellers in Rio or 1,076 in Sao Paulo).

Next, traveling costs are derived as the estimated driving times between pairs of locations, calculated by the C++ package of Project OSRM [38].

The demand of each order is set as the total order volume, calculated via the reported dimensions of each product in the order. 1% of the volumes are not specified in the data, and we replace them with the median demand. The vehicle capacity is set to 160 liters, which is about 10 times the average order demand. To avoid extreme outliers or packages larger than the whole capacity, we clip all the demands to a maximum of 100 liters. Finally, to avoid numerical precision issues, we convert all capacities and demands to milliliters ($\times$ $1000$) and round up to an integer.

The Sao Paulo dataset is visualized in Figure 2b. The process of data generation and problem instance sampling is illustrated in Figure 11.

**Optimization time budget**

To guarantee that each method is assigned the same time budget in total, the runtime of the initial solutions' generation is reduced from the solver budget. For example, a reported runtime of 60s in VRP-500 with EARLI, consists of 0.8s for the RL and 59.2s for the GA.

**Vehicle constraint**

In VRP literature, some solvers set the objective as minimization of the traveling cost, while others use a hierarchical objective: first minimize the number of vehicles, then the traveling cost. To allow a coherent comparison between solvers, we make both objectives equivalent in our setup. To that end, we set the vehicle budget as the minimal number of vehicles that still permits a known feasible solution. Hence, any feasible solution has the same number of vehicles, and the solution cost becomes the sole metric to evaluate feasible solutions.

**Statistical analysis and comparison of solution costs**

For each experimental configuration – problem size (number of customers) and data type (synthetic or real) – we used 256 test instances, generated i.i.d from the distribution of problem instances described above. All the solvers were tested on the same 256 problem instances. All test instances are generated with different seeds from those used to generate training instances. In real data, train and test instances are sampled from orders corresponding to different ranges of dates – one epoch for train instances and one for test instances.

Since most of the variance between solution costs comes from the problem instance itself, we normalize each solution cost as the gap from the best solution known to us: $gap = \frac{cost - best\ cost}{best\ cost}$. For every problem instance, the best solution is defined over all solvers, initialization schemes and time budgets in the experiment. When reporting the mean gap per solver and time budget, over the 256 test instances, we also report 95% confidence intervals for the mean, calculated via bootstrapping. Figure 5b also displays the complete set of 256 data points for certain configurations.

**Comparing costs when not all solutions are feasible**

Even in short time-budgets, all the experimented GAs succeed in finding a feasible solution in most of the test instances in both synthetic and realistic datasets (e.g., >95% after 2s for 500 customers, as detailed in the Extended Data). Still, this means that in a few problem instances, the solvers return an infeasible solution.

Whenever feasible solutions are missing, we calculate the average costs only over problem instances in which a feasible solution was found by *all* the methods in the figure. This guarantees that (a) only feasible costs are counted; (b) every two methods are compared on exactly the same set of problem instances. This holds for Figure 3, Figure 4, Figure 6 and Figure 7.

To compare solvers over the complete test set of 256 instances – including instances with no feasible solutions – we present Figure 5a. In Figure 5a, solvers are compared on each problem via the hierarchical metric: the first priority is solution feasibility, and the second is minimizing the cost.

**Baseline implementation details**

In the experiments above we compared EARLI to alternative initialization baselines, including greedy nearest neighbor, TNT, QBL, and DeepACO. The baseline implementation details under the VRP settings are available in the supplemental.

**Hardware**

All experiments presented in this paper were run on an Ubuntu machine with an NVIDIA A100 80GB Tensor Core GPU, and Dual AMD Rome 7742 with 16 cores.

To test the sensitivity of our experiments to the hardware, we also reproduced the main results on a different, local machine setup, as presented in the supplementary information. For that, we used an Ubuntu machine with an NVIDIA RTX A6000 GPU, and an AMD Ryzen 9 7950X 16-Core Processor.

## Data and code availability

Our complete Olist-based dataset is publicly available in GitLab. This includes both Sao Paulo and Rio data, separated into train, validation and test instances, as well as code to generate new problem instances from the raw orders data. For future research, we also publish in GitLab an interface for injecting initial solutions to HGS and PyVRP solvers.

## Autor contributions

I.G. and E.M. implemented the RL agent and the training procedure. I.G., P.S., H.L., R.G. and E.M. aided in interfacing the GAs and formatting the data accordingly. P.S., H.L. and R.G. helped developing the cuOpt GA. I.G. designed the Olist-based real-data benchmark. I.G. and E.M. analyzed the experimental results. I.G., E.M. and G.C. designed the visualization of the results and the figures. I.G. and E.M. wrote the original draft. S.M., A.F., G.C. and E.M. supervised the study. All authors aided in experimental design, interpretation of results, and critical revision of the manuscript.

# Extended Data

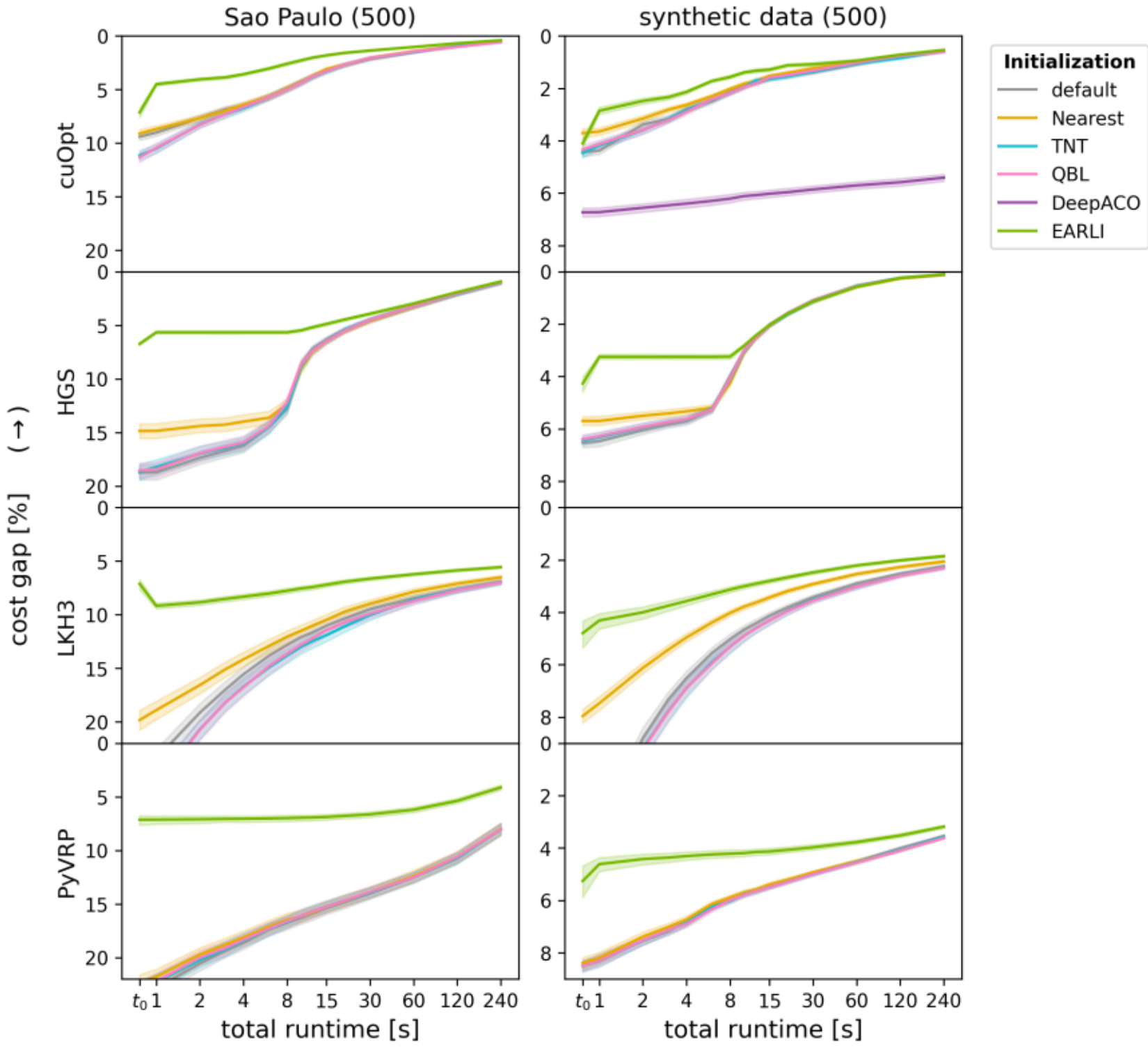

**Average cost gaps for 500 customers.** Each row presents the cost gaps of a different solver, across initialization methods and 256 test instances. DeepACO is a complete framework (initialization+solver) and thus is only shown once.

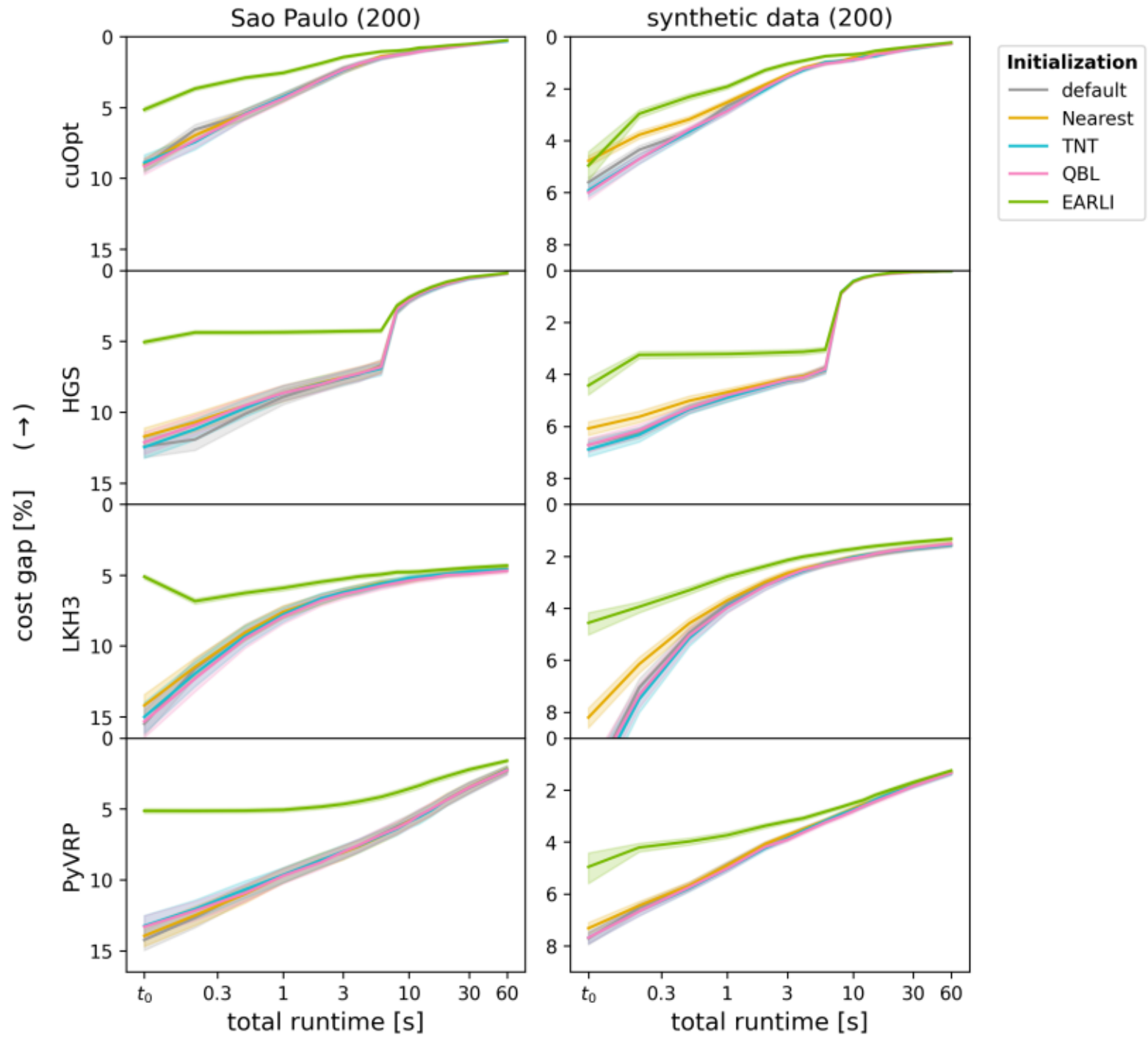

**Average cost gaps for 200 customers.**

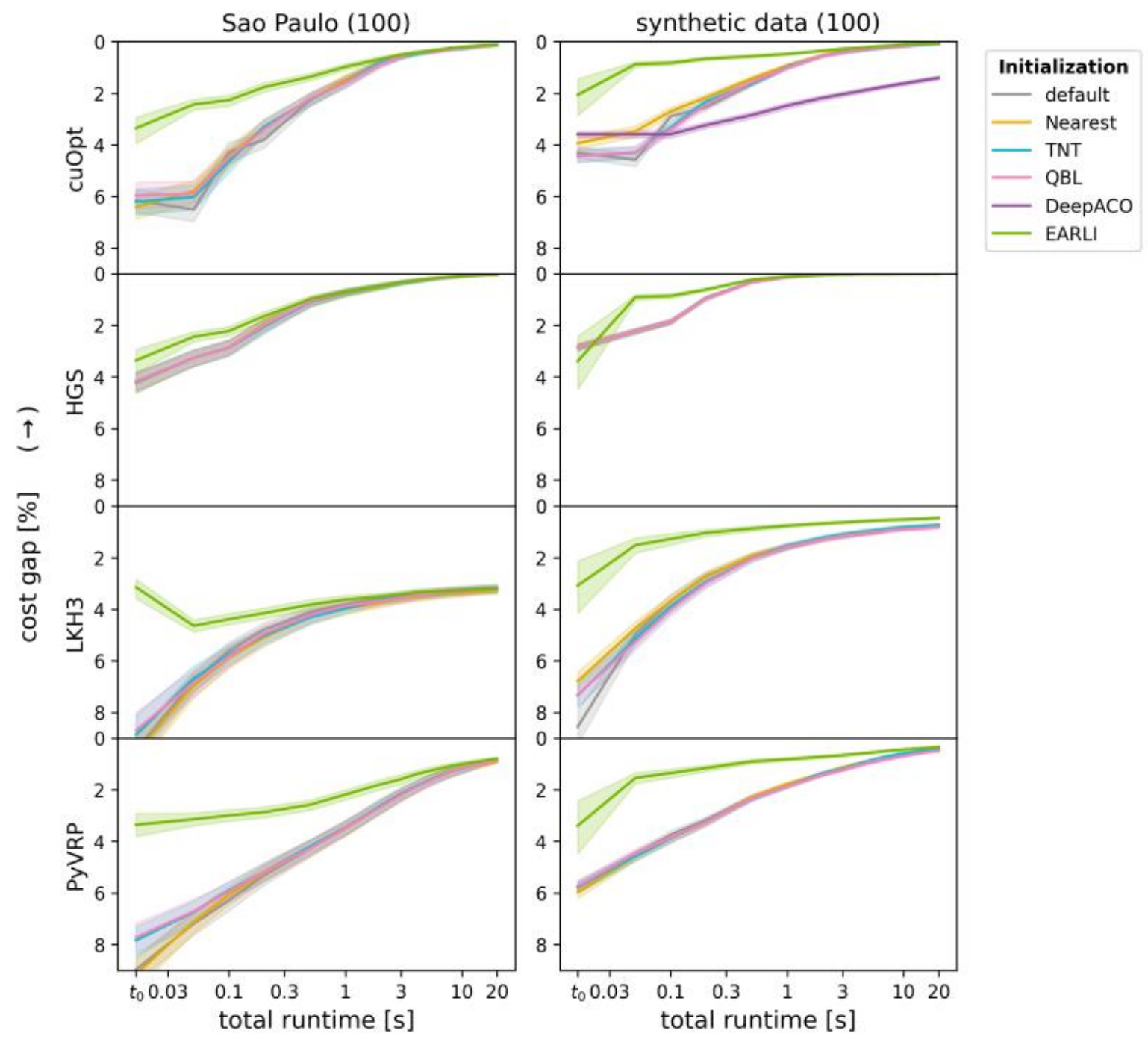

**Average cost gaps for 100 customers.**

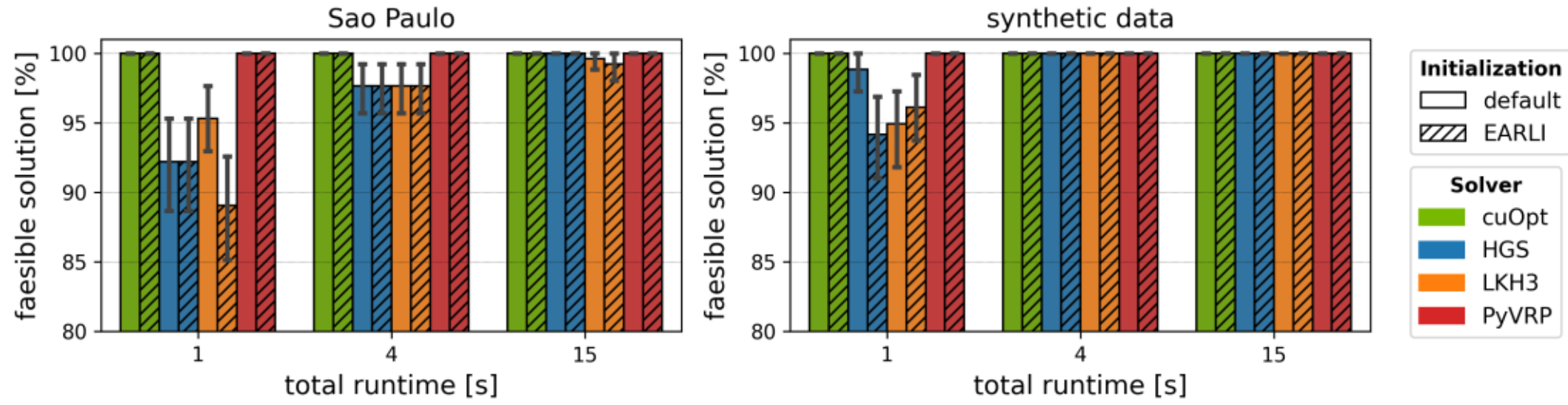

**Feasible solutions:** Percent of problem instances where the solver found a feasible solution, out of 256 test instances for 500 customers. Even in this problem size, all solvers usually find a feasible solution within a few seconds. Error bars correspond to 95% confidence intervals.

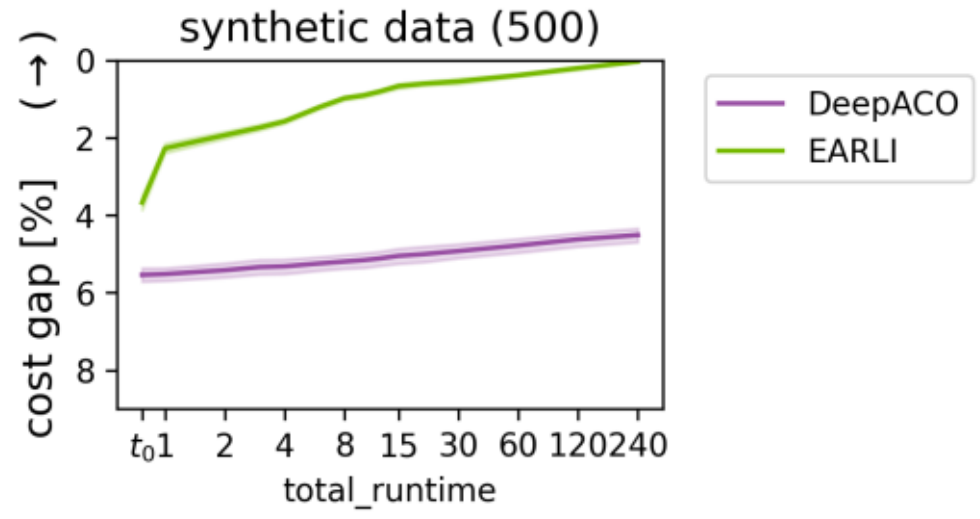

**DeepACO vs. EARLI** on VRP instances sampled from DeepACO's train distribution (500 customers, capacity 50). Despite the minor distribution shift for EARLI (which was trained with capacities 40-80), EARLI's advantage stands. In this experiment, EARLI is tested with the cuOpt solver.

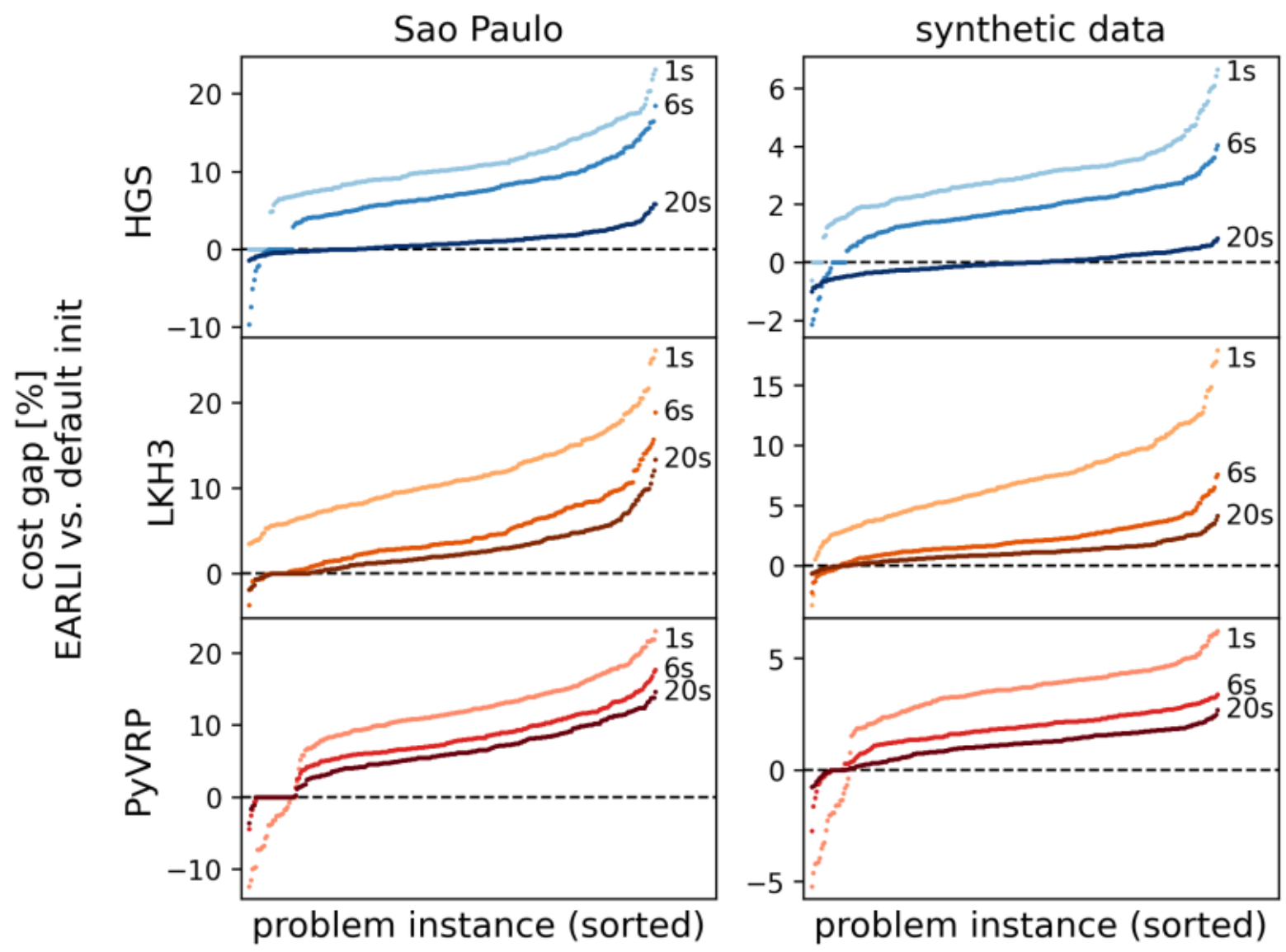

**Cost reduction of EARLI** (RL initialization vs. default initialization), for each of the 256 test instances with 500 customers, for time-budgets 1s, 6s, 20s. This is an extension of Figure 5b to the solvers HGS, LKH3 and PyVRP.

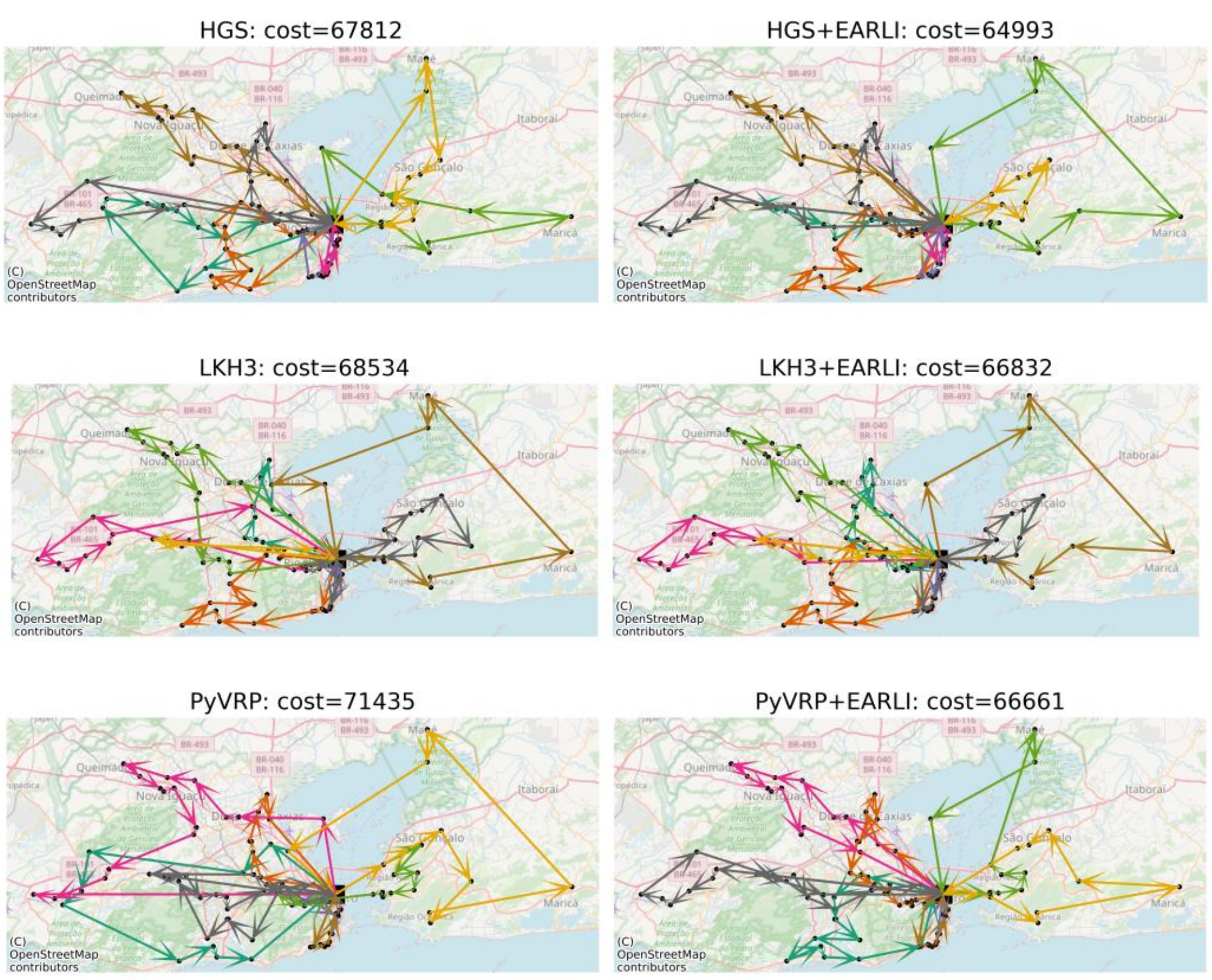

**Sample solutions visualization,** after 1s time-budget, with and without EARLI, on top of the solvers HGS (top), LKH3 (mid), and PyVRP (bottom). CuOpt is displayed in Figure 8.

# Supplementary Information

## Alternative hardware experiments

The experiments presented in this work are reported in terms of performance per wall-clock time-budget. Naturally, time-budget has a different meaning for different hardware. In this section, we demonstrate that our key message is robust to the hardware, by reproducing the main experiments with a different hardware.

Specifically, the main experiments were run on a cloud server with an NVIDIA A100 80GB Tensor Core GPU, and a Dual AMD Rome 7742 CPU with 16 cores. In this section, we use the hardware of a local machine: an NVIDIA RTX A6000 GPU, and an AMD Ryzen 9 7950X 16-Core Processor.

We reproduce the experiments for the Sao Paulo benchmark with 500 customers, for 64 problem instances, for cuOpt and HGS, up to a time budget of 120 seconds.

As displayed below, similarly to Figure 3, EARLI still improves the solution costs given a fixed time budget.

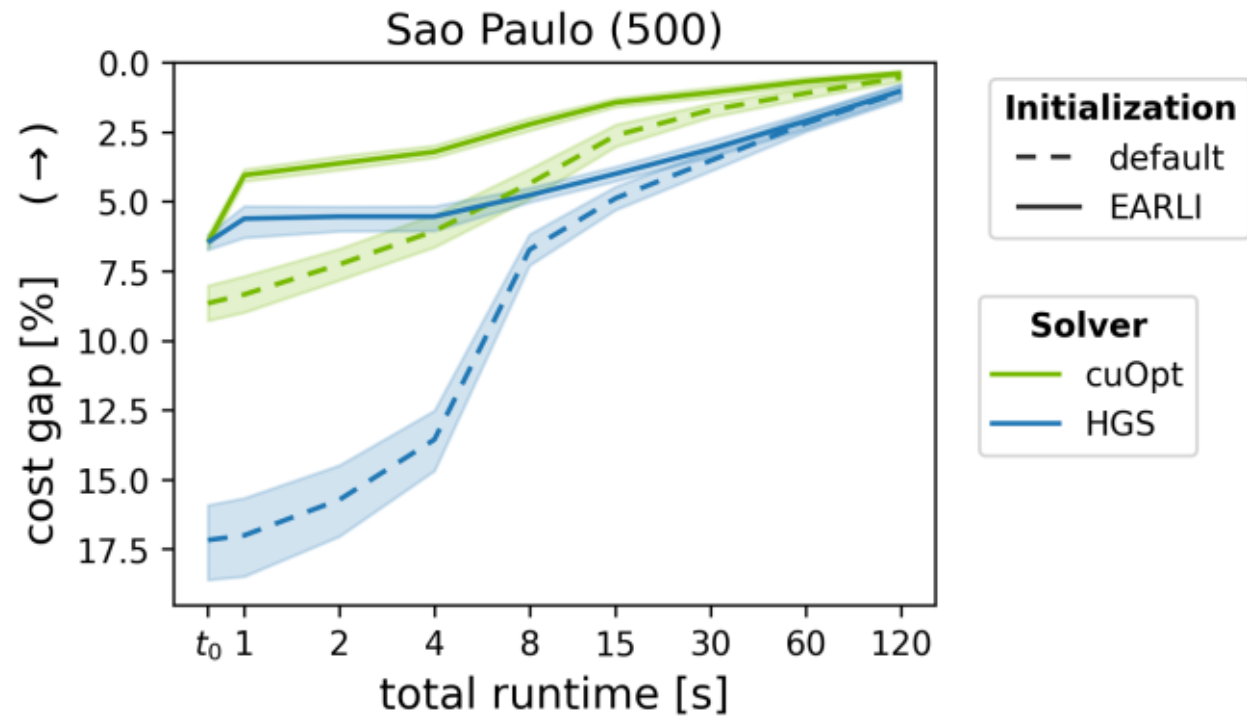

**Results on the alternative hardware of a local machine:** Average cost gaps for 500 customers in Sao Paulo benchmark.

## Test-time learning experiments

Previous studies applied ML optimization tools to combinatorial problems at inference time, to assist traditional solvers in the beginning [32] [33] [35] or throughout the search process [34]. Such test-time learning may aid in finding the best final solution, but is expensive whenever running time or computation are limited. In this section, we demonstrate that test-time learning may indeed be impractical for a complex problem with a short inference time.

To that end, we use a variant of our training method, designed to learn quickly how to solve one specific problem instance. This is obtained via the following modifications: (a) training repeatedly on the same problem instance; (b) for a small number of train epochs; (c) with re-tuned hyper-parameters to accelerate training (learning rate, episodes per train epoch, and gradient steps per train epoch).

Among all the hyper-parameter configurations, the table below displays the results of the two "best" configurations: the one with the fastest training, and the one with the best final costs. Evidently, even minutes of training were insufficient for learning a VRP instance for hundreds of customers without any prior experience. This demonstrates the need to leverage data in advance in order to solve complex combinatorial problems in short time budgets.

| #customers | Selected configuration | Configuration | Train time [s] | Cost gap |
|---|---|---|---|---|
| 100 | Fastest | learning_rate: 3e-4<br>epoch_length: 64<br>gradient_steps: 1 | 34 | 203% |
| | Best costs | learning_rate: 3e-4<br>epoch_length: 1024<br>gradient_steps: 8 | 1353 | 78% |
| 500 | Fastest | learning_rate: 3e-2<br>epoch_length: 64<br>gradient_steps: 8 | 197 | 361% |
| | Best costs | learning_rate: 3e-2<br>epoch_length: 256<br>gradient_steps: 8 | 4284 | 320% |

## Implementation details: solution sampling and solver initialization

At inference time, EARLI observes a batch of new problem instances and uses the trained RL agent to generate $N$ solutions per instance. The first solution is deterministic, obtained by always selecting the highest-probability action, corresponding to the maximum-likelihood solution under the learned policy. The remaining $N-1$ solutions are stochastic, sampled from the policy output distribution. In the experiments we use $N = 8$ initial solutions per problem instance, which typically suffices to provide at least one feasible solution (see Extended Data), while keeping a computational overhead low. Each solution is refined using the local-search operator of [44], and the resulting solutions are used to initialize the solver population. If the solver only permits fewer than $N$ solutions (LKH3), we choose the lowest-cost solutions among the $N$. If the solver's initial population is larger than $N$ solutions (HGS and PyVRP), we let the solver complete the population with its own internal random initialization, up to its standard initial population size.

## Implementation details: baselines

### TNT and QBL

Tent map (TNT) [49] and Quasi-opposition Based Learning (QBL) [48] are used to quickly generate solution samples that are both well-scattered and biased toward good fitness values. They apply a tent map and an opposition map, respectively, to a set of random solutions, and keep the best solutions out of the joint set of original and mapped solutions. Both methods were found particularly beneficial for GA initialization, compared to a variety of alternative initialization methods [28].

As both methods operate in a continuous real-valued space, we use random-key encoding [60] to convert samples to discrete visiting order (compatible with TSP), and then cut the tour of visits into capacity-feasible routes (compatible with VRP). The resulting solutions still use a high number of vehicles (routes); thus, to improve capacity utilization and minimize the number of vehicles, we add a post-processing procedure that fills each route with customers of later routes that fit the residual capacity. To eliminate any effect of implementation efficiency, we let each solver run for the whole time-budget after initialized by TNT or QBL – as if TNT and QBL themselves require no running time.

**DeepACO**

DeepACO [36] combines RL with Ant Colony Optimization (ACO) for VRP as follows: the RL agent learns to predict a score for each edge in the VRP graph, which is used as a prior heatmap for the ACO edge selection. Similarly to our work, the RL agent is trained to generalize from similar VRP instances. To test DeepACO on our test data, we use the official code repository of [36]. The repository already includes RL agents trained for VRP instances with 100 and 500 uniformly-distributed customers, similarly to our synthetic data. We run DeepACO with these RL agents on our corresponding test datasets. We modified DeepACO's interface to accept a time-budget instead of a number of iterations, and let it run iterations until exceeding the budget (such that the actual running time is at least as long as the budget). We notice that for 500 customers, DeepACO's train data corresponds to a vehicle capacity (50) different than ours (40-80); in the Extended Data we demonstrate that EARLI's advantage stands regardless of the capacity.